\documentclass{article}

\usepackage{PRIMEarxiv}
\usepackage[table,x11names]{xcolor}

\usepackage[utf8]{inputenc} 
\usepackage[T1]{fontenc}    
\usepackage{hyperref}       
\usepackage{url}            
\usepackage{booktabs}       
\usepackage{amsfonts}       
\usepackage{nicefrac}       
\usepackage{microtype}      
\usepackage{lipsum}
\usepackage{fancyhdr}       
\usepackage{graphicx}       
\usepackage{natbib}
\graphicspath{{media/}}     

\usepackage{hyperref}
\usepackage{url}
\usepackage{graphicx} 

\usepackage{booktabs}
\usepackage{inconsolata}
\usepackage{multirow}
\usepackage{amsmath}
\usepackage{natbib}

\usepackage{amsmath}  
\usepackage{amssymb}
\usepackage{newfloat}
\usepackage{listings}
\usepackage{wrapfig}

\usepackage{arydshln}
\usepackage{float}

\title{FineScope: SAE-guided Data Selection Enables Domain-Specific LLM Pruning \& Fine-Tuning
}
\author{
  Chaitali Bhattacharyya\textsuperscript{1},
  Hyunsei Lee\textsuperscript{2},
  Junyoung Lee\textsuperscript{1},
  Shinhyoung Jang\textsuperscript{2},
  Il Hong Suh\textsuperscript{3},
  Yeseong Kim\textsuperscript{1}, \\[4pt]
  \textsuperscript{1}\,POSTECH \quad
  \textsuperscript{2}\,DGIST \quad
  \textsuperscript{3}\,COGA Robotics
}

\begin{document}
\maketitle

\begin{abstract}
Large Language Models (LLMs) are typically trained on diverse, general-purpose datasets, enabling broad generalization but incurring substantial computational costs. However, real-world applications often require efficient models tailored to narrow, domain-specific tasks. In such settings, large model capacity and generality are unnecessary, and traditional fine-tuning pipelines struggle under resource constraints. We introduce FineScope, a framework that addresses this challenge by tightly coupling domain-aware data selection with model pruning and fine-tuning. Starting from a small set of user-provided seed examples, FineScope trains sparse autoencoders (SAEs) on intermediate model activations to automatically extract semantically aligned examples from large unlabeled corpora. The curated dataset then guides structured pruning to preserve domain-relevant substructures and supports self-distillation fine-tuning to recover task-specific performance. Experiments across STEM, humanities, social sciences, math, and coding domains show that FineScope consistently outperforms baseline fine-tuning approaches while enabling up to $35\%$ parameter pruning. On math reasoning tasks, it achieves an average improvement of ~11.50 points across pruned models. Code will be available.
\end{abstract}

\section{Introduction}\label{intro}
Large Language Models (LLMs) have demonstrated strong generalization across a wide range of tasks due to their training on broad and diverse datasets\citep{borealis2024highlevel, b1}. However, most real-world applications run in resource-constrained environments~\citep{upstagesolar, llama3, nvidia70} and require models to perform well on a narrow set of domain-specific tasks. In these cases, the broad capabilities of large models are often unnecessary, and much of their computational cost is spent on supporting functions that are irrelevant to the target domain. This creates a growing need for models that are both efficient and specialized for specific applications.
\begin{figure}[t]
    \centering
    \includegraphics[width=1.0\linewidth]{figures/fig1.pdf}
      \caption{Overall accuracy (with different model pruning ratio) on the STEM, Social Sciences, and Humanities domains after fine-tuning with  Self-Instruct dataset~\citep{selfinstruct} and FineScope dataset. Despite its smaller size, FineScope sustains higher accuracy under aggressive pruning, underscoring its data efficiency and quality.}
    \label{fig:init_overall}
\end{figure}
Despite significant progress in model compression techniques such as pruning and quantization, effective domain adaptation still depends heavily on the availability of high-quality, domain-specific data. In an ideal setting, a well-constructed dataset tailored to the target domain would be used for fine-tuning~\citep{csr, mcq, problemstatementpaper}. However, such datasets are rarely available in practice. Most existing approaches rely on general-purpose instruction datasets~\citep{alpaca} or manually curated corpora, which can be noisy, expensive to build, or misaligned with the intended application. As a result, models are often fine-tuned with suboptimal data which hinders performance recovery after compression.

A key challenge in domain adaptation is the limited availability of high quality, domain-aligned data for fine tuning. While much of the existing work focuses on model compression and adaptation techniques, these approaches often assume access to suitable data. In practice, the effectiveness of compressed models depends heavily on the quality and relevance of the fine tuning dataset~\citep{lima}. Without data that closely reflects the target domain, even the most advanced compression strategies are unlikely to maintain strong performance.

To address this challenge, we propose that data selection should be treated as a central part of the adaptation process. Inspired by findings that data quality can be more impactful than quantity~\citep{lima}, we argue that a small, carefully chosen subset of relevant data can support strong performance, even in heavily compressed models. Rather than treating data curation and model optimization as separate steps, we design an approach that connects them, allowing the data to guide both pruning and fine-tuning within a unified framework.

We present FineScope, a framework that automates domain-specific data selection and integrates it with model pruning and fine-tuning. FineScope begins with a small set of user-provided seed examples and uses a Sparse Autoencoder (SAE), trained on intermediate activations of a pretrained LLM, to identify semantically relevant samples~\citep{anthropy, kissane2024interpreting, yan2024encourage} from a large unlabeled corpus. These curated samples form a compact, high-quality dataset that reflects the target domain and is used to guide structured pruning. A modified self-distillation fine-tuning step then helps recover any task-relevant knowledge lost during compression.

FineScope enables the development of lightweight, domain-specialized models with minimal supervision and computational cost. As illustrated in Figure~\ref{fig:init_overall} the framework combines automatic data selection, structured pruning, and fine-tuning to produce compact models that retain domain-specific performance.
Across a range of domains and tasks, our experiments show that FineScope outperforms standard fine-tuning pipelines and significantly improves the performance of pruned models. These results demonstrate that integrating targeted data selection with model adaptation is a powerful and practical strategy for domain-specific LLM deployment in resource-constrained environments. Our contributions are as follows:

\begin{itemize}

\item We propose FineScope, a unified framework that connects domain-specific data selection with model pruning and fine-tuning to support efficient adaptation of large language models.
\item We introduce a novel use of Sparse Autoencoders trained on intermediate activations to identify semantically relevant data samples from large unlabeled corpora, starting from only a small seed set.
\item We develop a modified self-distillation fine-tuning approach that helps pruned models regain domain-relevant behaviors using the curated dataset.
\item We demonstrate that FineScope consistently improves performance over standard fine-tuning methods and enables effective domain adaptation, while significantly reducing model size.
    
\end{itemize}

\section{Related Work}\label{RW}
\textit{(1) Domain-Specific Language Models}
Recent efforts have adapted large language models to specialized domains by training or fine-tuning them on domain-specific datasets. Examples include PharmaGPT~\citep{pharmagpt} for biomedical applications, SaulLM~\citep{saullm} for legal tasks, Shai~\citep{shai} for asset management, BloombergGPT~\citep{bloomberg} for financial analysis, and MedPalm~\citep{medpalm} for medical question answering. Additional models such as ClimateBERT~\citep{climatebert}, ChatLaw~\citep{chatlaw}, and FinGPT~\citep{fingpt} focus on areas such as climate science, legal reasoning, and financial modeling.
While effective, these models typically rely on access to large, high-quality domain datasets and require full-scale retraining of billion-parameter models.
Few methods address the challenges of adapting models efficiently, particularly in settings with limited compute and annotated data. In contrast, our work focuses on enabling domain adaptation by automatically selecting relevant data from large unlabeled corpora and compressing models to reduce computational requirements without sacrificing task performance.

\textit{(2) Pruning}
Pruning is a widely used technique for reducing the size and computational cost of language models by removing parameters that have minimal impact on performance. Motivated by the lottery ticket hypothesis~\citep{lottery}, many methods aim to identify smaller subnetworks within large models that can be trained or fine-tuned to match the original model’s accuracy. Structured pruning, in particular, removes entire architectural components such as attention heads or feedforward blocks, resulting in models that are both compact and efficient in hardware implementation.

Existing pruning methods, both structured and unstructured~\citep{structured, taskstructured, other1, other2, other3, shearedllama}, are typically designed to maintain general-purpose capabilities and are applied independently of the data used for downstream fine-tuning. As a result, they may not fully account for the specific needs of domain adaptation.
In our approach, pruning is informed by data that is automatically selected for its relevance to a target domain. By integrating data selection into the pruning process, where we prune the model with respect to the domain-specific dataset, we aim to retain subnetworks that are more closely aligned with domain-specific behavior.

\textit{(3) Neural Representation Alignment}
While data-centric adaptation has gained increasing attention, most methods treat the model as fixed and focus solely on selecting or generating training examples. Instruction tuning and self-training approaches~\citep{lima, alpaca, selfinstruct} improve performance by using curated prompts or synthetic data but do not adapt the underlying model structure. Retrieval-based methods~\citep{bricken2023decomposing, gadre2023datacomp} select data based on surface-level similarity or metadata, which may not accurately reflect the model’s internal understanding of domain relevance.

Recent studies suggest that intermediate model activations encode meaningful signals related to both task characteristics and input semantics~\citep{anthropy}. However, this insight has rarely been applied to guide either data selection or model compression. Our work builds on this idea by using a Sparse Autoencoder trained on the top-$k$ intermediate activations of a pretrained model to estimate domain relevance. This allows us to identify training data that aligns with the model’s internal representations and use it to inform both pruning and fine-tuning.

To the best of our knowledge, FineScope is the first framework to jointly leverage latent model representations for both dataset construction and structured pruning, enabling efficient domain adaptation under resource constraints.

\begin{figure*}[t]
    \centering
    \includegraphics[width=0.85\linewidth]{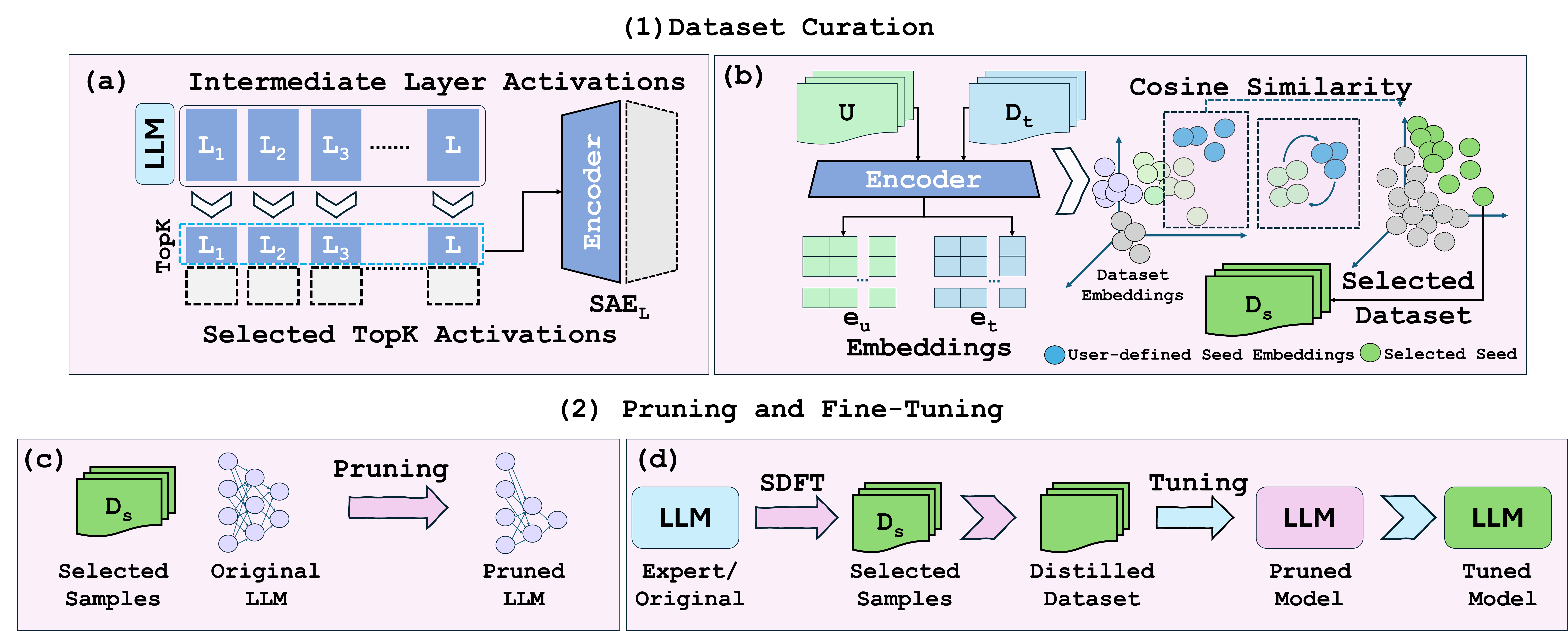}
    \caption{Overview of the FineScope: (1) \textit{Dataset Curation}: (a) Sparse Autoencoder (SAE) is trained on the top-$K$ activations and then used it to extract embedding from datasets, (b) Domain-specific dataset is curated by computing cosine similarity between target domain and the samples in the larger dataset. (2) \textit{Pruning and Fine-Tuning}: (c) Structured pruning is done w.r.t. selected dataset; (d) Fine-tuned the pruned model using modified self distillation. Here, $U$ denotes the larger dataset, while $D_t$ represents the target domain dataset. The corresponding embeddings are denoted by  $e_u$ and $e_t$ respectively.}
    \label{fig:overall}
\end{figure*}

\section{FineScope}\label{Method}
We present FineScope, a two-stage framework for efficient domain adaptation of large language models. The first stage selects domain-relevant examples from a large unlabeled corpus using a sparse autoencoder trained on the model’s internal activations. In the second stage, the selected data is used to guide structured pruning and fine-tuning through a modified self-distillation process. This approach enables the specialization of compact models that retain strong performance within the target domain.

\subsection{Methods}

\subsubsection{Training SAE}\label{sae_training}

A Sparse Autoencoder (SAE) is a neural network designed to learn compressed representations of input data while enforcing sparsity constraints on the hidden units~\citep{EleutherAI_SAE}. In our framework, SAEs serve as a mechanism for extracting domain-relevant features from pretrained LLM activations to identify domain-relevant samples from a large corpus. (We discuss why we used SAE encoder code as the embedding for retrieval in the supplement; see Section~\ref{sec:sup_sae}.)

Instead of operating directly on the raw model outputs, we train SAEs on activations from intermediate layers of the LLM, allowing us to capture a structured, low-dimensional representation of the underlying knowledge encoded in the model, as shown in Figure~\ref{fig:overall}a. Additionally, since processing all activations is computationally infeasible for a large corpus, our SAEs are adapted to learn only the representative activations in a way that highlights the most significant neurons, improving interpretability while discarding less relevant signals.

In decoder-only transformer models such as GPT-2 and LLaMa, the activation flow can be formally defined as follows~\citep{e2e}:
\begin{equation}\label{transformer}
   act^{(l)}(x) = f^{(l)}\left(act^{(l-1)}(x)\right), \text{for } l = 1, \dots, L-1.
\end{equation}
where $act^{(l)}(x)$ denotes the activations at layer $l$ given input $x$, and $f^{(l)}$ represents the transformation function at layer $l$, which typically includes multi-head self-attention, feed-forward operations, and residual connections. The final model output is computed as:
\begin{equation}\label{output}
   y = \text{softmax}\left(f^{(L)}\left(act^{(L-1)}(x)\right)\right).
\end{equation}

To train the SAE, we extract activations $act^{(\ell)}(x)$ from a selected layer of the pretrained LLM and feed them into the encoder network:
\begin{equation}\label{enc}
\mathrm{Enc}\!\left(act^{(\ell)}(x)\right) 
= \mathrm{ReLU}\!\left(W_e\,act^{(\ell)}(x) + b_e\right).
\end{equation}
The corresponding reconstruction from the decoder is given by:
\begin{equation}\label{sae}
\mathrm{SAE}\!\left(act^{(\ell)}(x)\right) 
= \mathrm{Dec}^\top \mathrm{Enc}\!\left(act^{(\ell)}(x)\right) + b_{\mathrm{dec}}.
\end{equation}

We train the SAE with a reconstruction loss plus an $\ell_1$ sparsity penalty on encoder activations:
\begin{equation}\label{eq:loss}
\mathcal{L}_{\mathrm{SAE}}
=\left\lVert \mathrm{SAE}\!\big(act^{(\ell)}(x)\big)-act^{(\ell)}(x)\right\rVert_2^2
+\lambda\left\lVert \mathrm{Enc}\!\big(act^{(\ell)}(x)\big)\right\rVert_1.
\end{equation}

The second term enforces sparsity by penalizing the activation magnitudes of the encoder. In our implementation, we use a scale-normalized reconstruction objective and additional stabilization terms; see the Supplementary Material for full loss details (Section~\ref{sec:sup_sae_objective}).

\subsubsection{Top-$K$ Activation Selection for SAE Training}\label{topk}

Training an SAE on the full activation width $d_\ell$ is costly. We therefore
preselect, for each layer $\ell$, a fixed subset of $K_{\text{act}} \ll d_\ell$
activation coordinates and train the SAE only on those coordinates.

Let $x$ be an input sequence, and let $e(x)\in\mathbb{R}^{T\times d_{\text{model}}}$
denote the embedding activations (treated as a differentiable input tensor; the
embedding parameters are frozen). Let $a^{(\ell)}(x)\in\mathbb{R}^{d_\ell}$ be a
sequence-level activation vector at a chosen hookpoint in layer $\ell$ . We define the per-coordinate input-sensitivity as
\begin{equation}
  s^{(\ell)}_j(x) \;=\;
  \left\|\frac{\partial a^{(\ell)}_j(x)}{\partial e(x)}\right\|_F,
  \label{eq:saliency_per_example}
\end{equation}
which corresponds to the Frobenius norm of the Jacobian row associated with
coordinate $j$.
In the supplement, we provide additional intuition for Jacobian-sensitivity as a coordinate-selection criterion (Section~\ref{sec:sup_jacob}), and an ablation comparing alternative Top-K selectors (Section~\ref{sec:sup_topk}).

\paragraph{Efficient estimation via randomized JVPs.}
Computing Eq.~\eqref{eq:saliency_per_example} exactly for all $j$ would require
access to the full Jacobian. Instead, we estimate the \emph{squared} saliency
using a Hutchinson-style estimator that only requires $R$ Jacobian vector
products (JVPs), with small $R$ (e.g., $R\in\{1,2,4\}$), independent of $d_\ell$.

Let $J^{(\ell)}(x) = \frac{\partial a^{(\ell)}(x)}{\partial \mathrm{vec}(e(x))}\in
\mathbb{R}^{d_\ell\times (T d_{\text{model}})}$, and sample probe vectors
$r^{(p)}\in\{\pm 1\}^{T d_{\text{model}}}$ with i.i.d.\ Rademacher entries.
For each probe, we compute a JVP
\begin{equation}
  u^{(p)}(x) \;=\; J^{(\ell)}(x)\,r^{(p)},
  \label{eq:jvp_def}
\end{equation}
which can be obtained by forward-mode automatic differentiation.
Then $\mathbb{E}\big[(u^{(p)}_j(x))^2\big] = \left\|J^{(\ell)}_{j,:}(x)\right\|_2^2$,
so we form the estimator
\begin{equation}
  \widehat{s}^{(\ell)}_j(x)^2
  \;=\;
  \frac{1}{R}\sum_{p=1}^R \left(u^{(p)}_j(x)\right)^2.
  \label{eq:hutchinson_diag}
\end{equation}

\paragraph{Global coordinate ranking.}
We aggregate over a reference corpus $D_0$:
\begin{equation}
  W^{(\ell)}_j
  \;=\;
  \mathbb{E}_{x\sim D_0}\big[\widehat{s}^{(\ell)}_j(x)\big]
  \;\approx\;
  \frac{1}{|D_0|}\sum_{x\in D_0}\widehat{s}^{(\ell)}_j(x),
  \label{eq:saliency_aggregate}
\end{equation}
and select a fixed index set
\begin{equation}
  I^{(\ell)} = \mathrm{TopK}\big(\{W^{(\ell)}_j\}_{j=1}^{d_\ell}, K_{\text{act}}\big).
  \label{eq:topk_indices_global}
\end{equation}

During SAE training we restrict activations to these coordinates:
\begin{equation}
  \tilde{a}^{(\ell)}(x) = a^{(\ell)}(x)\big[I^{(\ell)}\big]\in\mathbb{R}^{K_{\text{act}}},
  \label{eq:filtered_activation_main}
\end{equation}
and train a layer-specific SAE on $\tilde{a}^{(\ell)}(x)$.


\subsubsection{Dataset Curation}\label{datacurate}
After training, each SAE functions as a feature extractor to identify domain-relevant samples from a large unlabeled corpus. Starting with a small number of seed examples representative of the target domain (e.g., around ten samples), we aim to construct a curated subset $D_s \subseteq U$ from a broader, mixed-domain dataset $U$ by selecting examples that are most similar to the seed set in the learned embedding space. This selection process, which aligns samples in $U$ with the target domain using SAE-derived representations, is shown in Figure~\ref{fig:overall}b.

Using the trained SAE, we compute embeddings for all samples in both $D_t$ and $U$. The embeddings for the target domain are given by $ E_t = \{\text{SAE}(x) \mid x \in D_t\}$. Similarly, we compute embeddings for all samples in the larger dataset $U$, $E_U = \{\text{SAE}(x) \mid x \in U\}$.

To identify samples in $U$ that are most similar to $D_t$, we first compute SAE embeddings
$e_u = \mathrm{SAE}(x_u)$ for $x_u \in U$ and 
$e_t = \mathrm{SAE}(x_t)$ for $x_t \in D_t$. 
For each sample $x_u \in U$, we define its similarity to the target domain as

\begin{equation}
s(x_u) 
= \max_{x_t \in D_t} 
\mathrm{CosSim}\!\left(
\mathrm{SAE}(x_u), \mathrm{SAE}(x_t)
\right).
\label{eq:similarity}
\end{equation}

We then construct the curated dataset by selecting the Top-$M$ samples in $U$ with the highest similarity scores:

\begin{equation}
D_s 
= \operatorname{TopM}_{x_u \in U}
\big( s(x_u), M \big).
\label{eq:topk_selection}
\end{equation}

The final dataset $D_s$ contains samples from $U$ that are most semantically similar to the target domain $D_t$ based on the cosine similarity of their SAE embeddings. To ensure that $D_s$ remains a high-quality domain-specific dataset. In our evaluation, the value of $M$ across all SAEs is set to $100$ for consistency in selection.

\subsection{Pruning and Fine-Tuning} \label{pruningtuning}
\subsubsection{Pruning}
To enhance model efficiency while retaining domain-specific knowledge, we apply structured pruning using LLM-Pruner~\citep{llmpruner}. Rather than relying on general-purpose criteria, we guide the pruning process with a domain-specific dataset $D_s$ (as shown in Figure~\ref{fig:overall}c), so that only components relevant and active to the target domain are preserved. 
Guided by the domain-specific dataset $D_s$, the contribution of each model block is estimated using gradient-based attribution. Specifically, we compute an importance score for each block based on the first-order gradient of the task-specific loss with respect to $D_s$. This allows the pruning process to preserve components that are most critical for performance in the target domain.
(We provide an optimization-based interpretation of domain-conditioned pruning and why conditioning on $D_s$  preserves domain-critical components in the supplement; see Section~\ref{sec:sup_domain}).

We apply block-wise structured pruning using LLM-Pruner~\citep{llmpruner}, which removes redundant components of the model and significantly reduces inference cost. (We compare opther pruning methods under matched pruning ratios in the supplement Section~\ref{sec:sup_pruning}.)
The original model, denoted by $\mathcal{M}$, is pruned with respect to the domain-specific dataset $D_s$ and a pruning ratio $r$, resulting in a compressed model $\mathcal{M}_r$:

\begin{equation}\label{eq:pruning}
\mathcal{M}_r = \text{LLMPrune}(\mathcal{M}, D_s, r)
\end{equation}
Here, $r$ serves as a hyperparameter that controls the trade-off between model compactness and domain-specific performance.

\subsubsection{Fine-Tuning via Teacher-Guided Distillation (TGD)}

To mitigate the loss of domain-specific knowledge caused by pruning, 
we adopt a teacher-guided distillation procedure during fine-tuning. 
Unlike the original formulation of Self-Distillation Fine-Tuning (SDFT)~\citep{sdft}, 
which conditions on intermediate teacher (self-generated) outputs, 
our approach performs cross-model/unpruned original model knowledge transfer from a teacher model to the pruned student.

In our framework, we adapt SDFT to further refine the pruned model. We generate a distilled dataset using 
either the original model or a pretrained state-of-the-art teacher, and use it to fine-tune the pruned model. 
This process helps preserve domain-specific knowledge while improving generalization.
In the original SDFT formulation~\citep{sdft}, given an input $x$, context $c^t$, and output $y^t$ from the teacher model, the distilled output $y'$ is sampled as:
\begin{equation}\label{sdft}
    y' \sim f(y \mid c^t, x^t, y^t).
\end{equation}

We adapt this procedure in FineScope by modifying the fine-tuning objective for the pruned model. The resulting distillation loss is defined as:
\begin{equation}\label{mod_sdft}
    L_{\text{tgd}} = -\log f_{p}(y' \mid c^t, y^t),
\end{equation}
where $f_p$ denotes the pruned model and $L_{\text{tgd}}$ represents the distillation loss.

This teacher-guided distillation serves two purposes: (1) \textit{Knowledge Recovery}
    It transfers domain-specific knowledge that may have been weakened or removed during pruning. (2) \textit{Regularization}
    It reduces overfitting to the small curated dataset by exposing the pruned model to high-confidence teacher outputs.

Importantly, the teacher model is used only during distilled dataset construction. 
The student is trained independently thereafter without direct access to teacher logits or intermediate representations.

\section{Evaluation}
\noindent\textit{\textit{(1) Models}}. We evaluate our method using three models: Vicuna-7B~\citep{vicuna}, a fine-tuned version of LLaMa 2~\citep{llama2}; MathCoder-CL-7B~\citep{mathcoder}, a CodeLlama~\citep{codellama} variant; and LLaMa 3.1-8B~\citep{llama3}.

\noindent\textit{\textit{(2) Baselines}}.
We compare FineScope against six baseline settings:
(a) fine tuning the pruned model using randomly selected data of the same size,
(b) fine tuning with the full dataset containing mixed domains,
(c) fine tuning with Alpaca data using FineScope’s pruning strategy,
(d) evaluating pretrained models without any fine tuning,
(e) evaluating pretrained models fine tuned with FineScope-curated data, and
(f) comparisons against GPT-3 (6.7B and 175B)\citep{gpt3} and OLMO-7B\citep{olmo} (Table~\ref{tab:model_results}), along with GPT-3 (13B and 175B) (Table~\ref{tab:math})

\noindent\textit{\textit{(3) Tuning Tasks}.} 
We assess FineScope on three main tasks:

\noindent\textit{(a) Domain Specific Tuning.}
We prune models using domain specific datasets curated with our SAE-guided framework. The SAEs are trained on the RedPajama dataset~\citep{together2023redpajama}, which includes content from CommonCrawl, C4, GitHub, Wikipedia, Books3, ArXiv, and StackExchange, providing broad domain coverage. Using these SAEs, we curate domain specific subsets from OpenInstruct~\citep{haku}, which aggregates instruction datasets such as Alpaca~\citep{albertying}, Self Instruct~\citep{selfinstruct}, GPT 4 Instruct, Roleplay~\citep{GPTeacher}, Code Alpaca~\citep{codealpaca}, and Dolly~\citep{DatabricksBlog2023DollyV2}. Based on user provided seed samples, we extract curated datasets for STEM (2,100 samples), Social Sciences (2,401 samples), and Humanities (2,374 samples), selecting the most frequently chosen samples across all trained SAEs.

\noindent\textit{(b) Subdomain Specific Tuning (Math).}
To assess FineScope’s effectiveness at a more granular level, we evaluate fine tuning on mathematical subdomains. SAEs are trained on the MetaMath dataset~\citep{metamath}, and used to curate subsets from the Math dataset~\citep{hendrycksmath2021}. In this setting, subdomains are merged into a unified pool for curation. From this pool, we extract the Pre Algebra, Algebra, and Counting and Probability subsets. Models are fine tuned using OpenMath2~\citep{openmath}, and Notus 7B~\citep{notus} serves as the Alpaca tuned baseline. Evaluations are conducted separately on each subdomain test set (Table~\ref{tab:math}).

\noindent\textit{(c) Coding Specific Tuning.}
Following the same SAE guided curation approach, we construct a code focused dataset from OpenInstruct, resulting in 1,200 examples. To evaluate model performance on code generation, we fine tune on this curated dataset and assess results using the HumanEval~\citep{humaneval} and MBPP~\citep{mbpp} benchmarks (Table~\ref{tab:coding_only}).

\noindent\textit{(4) SAE\footnote{We used Anthropic style architecture~\citep{EleutherAI_SAE}.} training:} We ran SAE training~\citep{EleutherAI_SAE} using the AdamW optimizer with a learning rate of 1e-5, a batch size of $8$, and a Top-K value of $128$. We used GPT-4~\citep{gpt4} to generate the $10$ user-defined seeds.

\noindent\textit{(5) Finetuning:} We fine-tuned LMs (LORA-fine tuning~\citep{lora}) 
using the same AdamW optimizer at a 5e-5 learning rate, 
a batch size of $128$, lora rank of 32, and a $256$ cut-off length. For generating the distilled 
dataset, we have used the corresponding unpruned model.

\subsection{Experimental Results}\label{results} 
\begin{table}[t]
  \caption{Performance comparison of FineScope-tuned models versus baselines across STEM, Social Sciences (Social Sci.), and Humanities (Hum.) domains.}
  \vspace{1em}
  \centering
  \scriptsize
  \setlength{\tabcolsep}{1pt}
  \renewcommand{\arraystretch}{0.1}
  \begin{tabular}{l l l l c c c}
    \toprule
    Model & Pruned & Tuned & Dataset & STEM & Social Sci. & Hum. \\
    \midrule
    \multirow{7}{*}{Vicuna~\citep{vicuna}} 
      & $\times$     & $\times$     & --        & 33.10 & 40.23 & 43.69 \\
      & $\checkmark$ & $\times$     & --        & 17.17 & 20.11 & 20.80 \\
      & $\checkmark$ & $\times$     & Random    & 18.52 & 21.29 & 20.21 \\
      & $\checkmark$ & $\times$     & Full-OI   & 29.09 & 35.43 & 36.19 \\
      & $\checkmark$ & $\times$     & Alpaca    & 30.61 & 35.44 & 36.11 \\
      & $\times$     & $\checkmark$ & FineScope & \underline{33.32} & \underline{40.21} & \underline{42.43} \\
      \rowcolor{blue!13}
      & $\checkmark$ & $\checkmark$ & FineScope & \textbf{31.12} & \textbf{36.23} & \textbf{36.55} \\
    
    \midrule
    
    \multirow{7}{*}{MathCoder-CL~\cite{mathcoder}}
      & $\times$     & $\times$     & --        & 31.14 & 11.11 & 9.22  \\
      & $\checkmark$ & $\times$     & --        & 13.32 & 8.02  & 3.67  \\
      & $\checkmark$ & $\times$     & Random    & 12.94 & 7.53  & 4.59  \\
      & $\checkmark$ & $\times$     & Full-OI   & 23.91 & 12.81 & 12.67 \\
      & $\checkmark$ & $\times$     & Alpaca    & 25.14 & 13.11 & 12.33 \\
      & $\times$     & $\checkmark$ & FineScope & \underline{34.96} & \underline{32.91} & \underline{31.66} \\
      \rowcolor{blue!13}
      & $\checkmark$ & $\checkmark$ & FineScope & \textbf{25.89} & \textbf{13.81} & \textbf{13.68} \\
    
    \midrule
    
    \multirow{7}{*}{LLaMa3.1~\citep{llama3}}
      & $\times$     & $\times$     & --        & 48.01 & 49.61 & 49.32 \\
      & $\checkmark$ & $\times$     & --        & 30.59 & 31.33 & 33.62 \\
      & $\checkmark$ & $\times$     & Random    & 29.04 & 30.93 & 33.71 \\
      & $\checkmark$ & $\times$     & Full-OI   & 39.32 & 39.91 & 40.93 \\
      & $\checkmark$ & $\times$     & Alpaca    & 38.22 & 40.19 & 39.79 \\
      & $\times$     & $\checkmark$ & FineScope & \underline{48.84} & \underline{51.66} & \underline{51.45} \\
      \rowcolor{blue!13}
      & $\checkmark$ & $\checkmark$ & FineScope & \textbf{40.55} & \textbf{41.07} & \textbf{41.19} \\
    
    \midrule
    
    GPT-3 (6.7B)   & $\times$ & $\times$ & -- & 35.10 & 49.20 & 42.10 \\
    OLMO           & $\times$ & $\times$ & -- & 22.19 & 31.01 & 30.26 \\
    GPT-3 (175B)   & $\times$ & $\times$ & -- & 36.70 & 50.40 & 40.80 \\
    
    \bottomrule
  \end{tabular}
  \label{tab:model_results}
\end{table}
Table~\ref{tab:model_results} presents evaluation results on the MMLU dataset across three domains. Models adapted using FineScope, through pruning and fine tuning with SAE-curated domain specific data, achieve average performance gains of 3.8\% over Alpaca tuning and 4.45\% over OpenInstruct (Full OI) across all domains and model types. Among the three evaluated models, MathCoder CL shows the most significant improvement, with gains of 8.28\% in STEM, 7.8\% in Social Sciences, and 7.9\% in Humanities. These results indicate that SAE-guided data selection not only improves domain adaptation but also enables pruned models to recover performance that would otherwise be lost under aggressive compression. Despite using fewer data points, our method outperforms OpenInstruct, underscoring the importance of data quality and domain alignment over quantity alone.

Pruning without domain guidance results in substantial performance degradation, reaching up to 50.17\% on average for Vicuna across all domains. LLaMa 3.1 shows the smallest drop, likely due to its more balanced initial performance and the ability of domain focused pruning to retain essential parameters. Compared to GPT-3 (6.7B and 175B) and OLMO 7B, our pruned models, with approximately 30\% fewer parameters, outperform in most settings. GPT-3 achieves stronger results in Social Sciences, and the 175B variant exceeds our models in Humanities, but FineScope tuned models consistently outperform OLMO 7B across all domains.

\begin{table}[t]
  \centering
  \caption{Performance comparison of FineScope-tuned models versus
   baselines across Pre-algebra (Pre-alg.), Algebra (Alg.), and Counting and 
   Probability (C.\&P.) domains.}
  \scriptsize
  \setlength{\tabcolsep}{0.4pt}
  \renewcommand{\arraystretch}{0.6}
  \begin{tabular}{l l l l c c c}
    \toprule
    Model & Pruned & Tuned & Dataset & Pre-alg. & Alg. & C.\&P. \\
    \midrule

    \multirow{7}{*}{Vicuna~\citep{vicuna}}
      & $\times$     & $\times$     & --        & 14.31 & 10.17 & 8.11 \\
      & $\checkmark$ & $\times$     & --        & 0.11  & 0.00  & 0.00 \\
      & $\checkmark$ & $\times$     & Random    & 0.00  & 0.00  & 0.00 \\
      & $\checkmark$ & $\times$     & Full-Math & 12.73 & 8.91  & 5.48 \\
      & $\checkmark$ & $\checkmark$ & Alpaca    & 5.56  & 0.30  & 0.21 \\
      & $\times$     & $\checkmark$ & FineScope & \underline{15.46} & \underline{13.33} & \underline{10.43} \\
      \rowcolor{blue!13}
      & $\checkmark$ & $\checkmark$ & FineScope & \textbf{12.91} & \textbf{10.12} & \textbf{7.01} \\

    \midrule

    \multirow{7}{*}{MathCoder-CL~\citep{mathcoder}}
      & $\times$     & $\times$     & --        & 11.60 & 16.77 & 13.38 \\
      & $\checkmark$ & $\times$     & --        & 0.59  & 2.33  & 0.29 \\
      & $\checkmark$ & $\times$     & Random    & 0.00  & 0.00  & 0.00 \\
      & $\checkmark$ & $\times$     & Full-Math & 9.01  & 12.72 & 10.05 \\
      & $\checkmark$ & $\checkmark$ & Alpaca    & 1.29  & 6.94  & 3.33 \\
      & $\times$     & $\checkmark$ & FineScope & \underline{14.73} & \underline{17.75} & \underline{15.43} \\
      \rowcolor{blue!13}
      & $\checkmark$ & $\checkmark$ & FineScope & \textbf{10.54} & \textbf{15.51} & \textbf{11.64} \\

    \midrule

    \multirow{7}{*}{LLaMa3.1~\citep{llama3}}
      & $\times$     & $\times$     & --        & 32.77 & 29.87 & 20.35 \\
      & $\checkmark$ & $\times$     & --        & 11.41 & 7.99  & 5.01 \\
      & $\checkmark$ & $\times$     & Random    & 7.04  & 8.01  & 6.93 \\
      & $\checkmark$ & $\times$     & Full-Math & 30.72 & 31.67 & 18.34 \\
      & $\checkmark$ & $\checkmark$ & Alpaca    & 9.23  & 5.56  & 9.10 \\
      & $\times$     & $\checkmark$ & FineScope & \underline{34.46} & \underline{31.85} & \underline{23.18} \\
      \rowcolor{blue!13}
      & $\checkmark$ & $\checkmark$ & FineScope & \textbf{30.83} & \textbf{32.21} & \textbf{19.34} \\

    \midrule

    GPT-3 (13B)~\citep{gpt3}  & $\times$ & $\times$ & -- & 6.80 & 5.30 & 4.50 \\
    GPT-3 (175B)~\citep{gpt3} & $\times$ & $\times$ & -- & 7.70 & 6.00 & 4.70 \\

    \bottomrule
  \end{tabular}
  \label{tab:math}
\end{table}

\begin{table}[t]
\centering
\caption{Performance comparison of FineScope-tuned models versus baselines across MBPP and HumanEval coding datasets.}
\scriptsize
\setlength{\tabcolsep}{4pt}
\renewcommand{\arraystretch}{0.8}
\begin{tabular}{l l l l c c}
\toprule
Model & Pruned & Tuned & Dataset & HumanEval & MBPP \\
\midrule

\multirow{7}{*}{Vicuna}  
    & $\times$      & $\times$      & --        & 0.14 & 0.03 \\
    & $\checkmark$  & $\times$      & --        & 0.04 & 0.00 \\
    & $\checkmark$  & $\times$      & Random    & 0.03 & 0.00 \\
    & $\checkmark$  & $\times$      & Full-OI   & 0.09 & 0.05 \\
    & $\checkmark$  & $\checkmark$  & Alpaca    & 0.07 & 0.00 \\
    & $\times$      & $\checkmark$  & FineScope & \underline{0.21} & \underline{0.13} \\
    \rowcolor{blue!13}
    & $\checkmark$  & $\checkmark$  & FineScope & \textbf{0.13} & \textbf{0.10} \\

\midrule

\multirow{7}{*}{MathCoder-CL}  
    & $\times$      & $\times$      & --        & 0.03 & 0.01 \\
    & $\checkmark$  & $\times$      & --        & 0.00 & 0.00 \\
    & $\checkmark$  & $\times$      & Random    & 0.00 & 0.00 \\
    & $\checkmark$  & $\times$      & Full-OI   & 0.10 & 0.09 \\
    & $\checkmark$  & $\checkmark$  & Alpaca    & 0.08 & 0.05 \\
    & $\times$      & $\checkmark$  & FineScope & \underline{0.20} & \underline{0.14} \\
    \rowcolor{blue!13}
    & $\checkmark$  & $\checkmark$  & FineScope & \textbf{0.11} & \textbf{0.10} \\

\midrule

\multirow{7}{*}{LLaMa3.1}  
    & $\times$      & $\times$      & --        & 0.50 & 0.46 \\
    & $\checkmark$  & $\times$      & --        & 0.26 & 0.13 \\
    & $\checkmark$  & $\times$      & Random    & 0.20 & 0.09 \\
    & $\checkmark$  & $\times$      & Full-OI   & 0.30 & 0.29 \\
    & $\checkmark$  & $\checkmark$  & Alpaca    & 0.25 & 0.13 \\
    & $\times$      & $\checkmark$  & FineScope & \underline{0.55} & \underline{0.48} \\
    \rowcolor{blue!13}
    & $\checkmark$  & $\checkmark$  & FineScope & \textbf{0.49} & \textbf{0.43} \\

\bottomrule
\end{tabular}
\label{tab:coding_only}
\end{table}

Table~\ref{tab:math} highlights significant performance improvements in math domains using our domain-specific tuning: Vicuna (+7.01), MathCoder-CL (+7.71), and LLaMa 3.1 (+18.45) versus Alpaca-tuned baselines. A similar trend is observed when finetuned with Math's full corpus (e.g.,+1.97 average performance gain for MathCoder-CL when compared with Full-Math corpus). However, pruning severely degrades Vicuna and MathCoder-CL's performance and Alpaca’s general-purpose instructions fail to restore performance due to a lack of semantic focus. Compared to GPT models, our tuned models achieve competitive performance, with differences likely due to GPT’s imbalanced training data limiting generalization. Despite reducing model size to approximately 71\% of the original, FineScope is able to restore performance by fine tuning on a semantically focused dataset.

As shown in Table~\ref{tab:coding_only}, our domain specific tuning dataset, FineScope, substantially improves coding performance on the HumanEval and MBPP benchmarks, especially after model pruning. When applied to pruned models, FineScope yields coding gains of +0.08 for Vicuna, +0.04 for MathCoder CL, and +0.27 for LLaMa 3.1 8B compared to tuning with the full Alpaca dataset. In all three models, pruning alone leads to a significant decline in code generation performance, and tuning on the Alpaca dataset fails to recover the loss.In comparison to OpenInstruct's full corpus (Full OI), FineScope delivers highly competitive and often superior results. For instance, the pruned LLaMa 3.1 model fine tuned with FineScope achieves scores of 0.49 on HumanEval and 0.43 on MBPP, outperforming the same model tuned with Full OI, which reaches 0.30 and 0.29, respectively. These results demonstrate that FineScope is more effective at producing specialized, high performing models than simply relying on a large general purpose corpus for tuning and pruning guidance.

\subsection{Effect of Teacher-Guided Distillation (TGD) and Pruning Dataset}

\textit{Teacher-Guided Distillation (TGD)} Table~\ref{tab:msdft_pruning} shows a consistent
 improvement in performance when modified SDFT is applied, 
 compared to standard fine tuning. 
 Across all three domains, models fine tuned with TGD outperform their counterparts. 
 For example, we observe performance gains of 1.9\% in STEM, 5.8\% in Social Sciences, 
 and 10.26\% in Humanities. This consistent improvement across diverse domains highlights 
 the effectiveness of a distillation based approach for enhancing model performance in 
 domain specific adaptation.

\textit{Pruning Dataset} Table~\ref{tab:msdft_pruning} highlights pruning with 
the FineScope domain aligned dataset, followed by fine tuning on the same domain data,
 results in an average accuracy gain of 11.8\% compared to pruning with a general purpose 
 corpus~\citep{bookcorpus}. This demonstrates that incorporating domain specific samples 
 during the pruning stage helps retain critical representations that are often lost when 
 using generic data, leading to consistently higher performance across STEM, 
 Social Science, and Humanities benchmarks.

\begin{table}[h!]  
\caption{Effect of Teacher-Guided Distillation(TGD) and domain-specific pruning method across domains.}
  \vspace{1em}
\centering
\scriptsize
\setlength{\tabcolsep}{3pt}
\renewcommand{\arraystretch}{0.9}
\begin{tabular}{lcccc}
\toprule
\multirow{2}{*}{Domain} & \multicolumn{2}{c}{TGD} & \multicolumn{2}{c}{Pruning} \\
\cmidrule(lr){2-3} \cmidrule(lr){4-5}
 & W/O & W/ & FineScope & Bookcorpus \\
\midrule
STEM            & 30.54 & \textbf{31.12} & \textbf{31.12} & 28.64 \\
Social Sciences & 34.25 & \textbf{36.23} & \textbf{36.23} & 33.24 \\
Humanities      & 33.15 & \textbf{36.55} & \textbf{36.55} & 31.03 \\
\bottomrule
\end{tabular}
\label{tab:msdft_pruning}
\end{table}

\subsection{Effect of TopK on Computation}
\begin{figure}[t]
    \centering
    \includegraphics[width=0.9\linewidth]{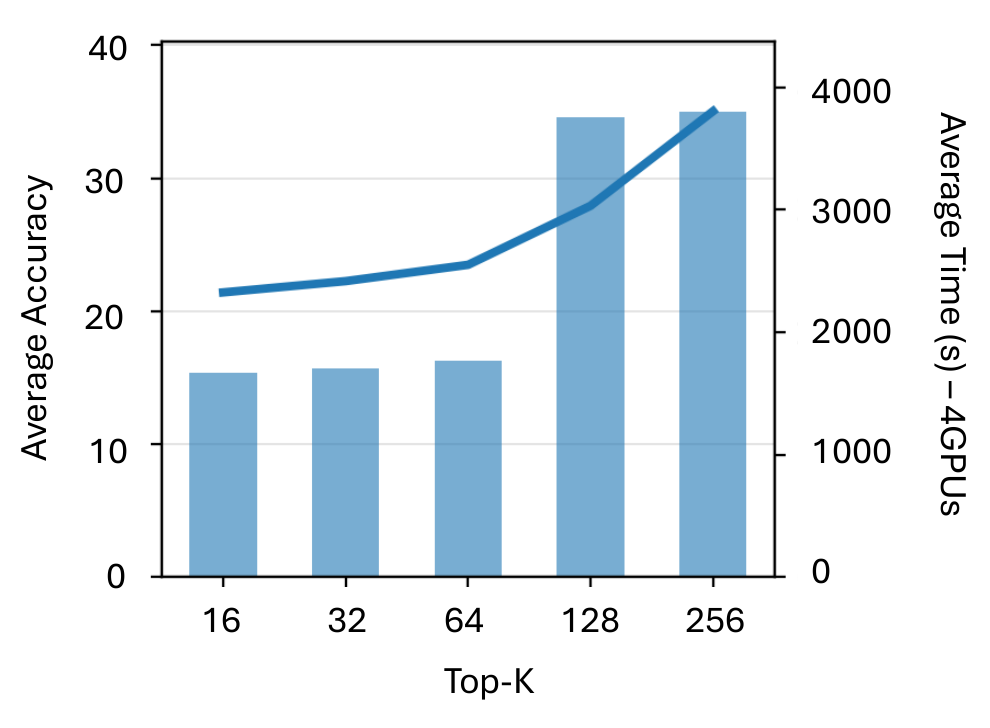}
    \caption{Impact of varying TopK on SAE's average reconstruction loss, average accuracy and training time for all transformer blocks.}
    \label{fig:sae_train}
\end{figure}
The choice of the hyperparameter $K$ for our Sparse Autoencoders (SAEs) reflects a trade-off between training efficiency and performance across domains. As shown in Figure~\ref{fig:sae_train}, increasing $K$ improves average accuracy on STEM, Social Sciences, and Humanities tasks but also leads to longer training times across all transformer blocks of the Vicuna 7B model. We find that setting $K$=128 provides a favorable balance, yielding strong average accuracy while keeping training time tractable. Further increasing $K$ to 256 offers only marginal accuracy gains at the cost of significantly higher computational overhead.

\subsection{Comparison with Synthetic Dataset}
\begin{figure}[t]
    \centering
    \includegraphics[width=1.0\linewidth]{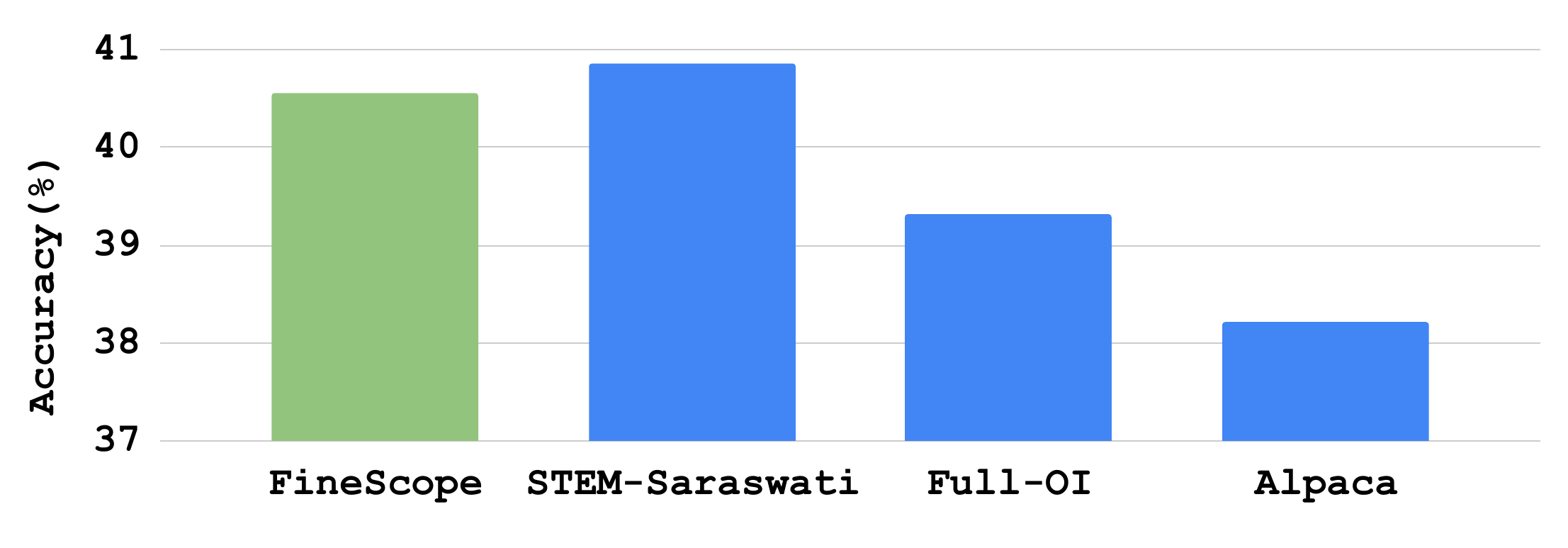}
    \caption{Performance comparison with synthetic STEM dataset.}
    \label{fig:placeholder}
\end{figure}

In Figrue~\ref{fig:placeholder} we evaluate our curated 
FineScope dataset (2.1K samples) against the publicly available 
synthetic STEM-Saraswati dataset~\citep{saraswati}, generated using 
GPT-4~\citep{gpt4}, as well as general-purpose finetuning datasets such 
as Alpaca and OpenInstruct. scope achieves comparable accuracy to 
STEM-Saraswati on STEM-specific tasks, demonstrating the high quality of our curated dataset. Moreover, when contrasted with general-purpose datasets, both STEM-Saraswati and FineScope achieve substantially higher performance, highlighting the critical role of high-quality, domain-specific data in enhancing model capabilities within specialized scientific domains.
\subsection{Effect of SAE-Embedding}
As shown in Table~\ref{tab:model_results_emb}, the method of data 
curation has a significant impact on model performance after fine-tuning. 
The results consistently demonstrate that curating the fine-tuning dataset 
using our interpretable SAE-embeddings leads to substantially better outcomes 
than using standard raw embeddings. To analyze the effect of the embedding 
type, we compare the performance of original 
and pruned models fine-tuned on datasets curated by each method. 
In addition to BERT-based embeddings, we evaluate a modern sentence 
embedding model (\texttt{all-mpnet-base-v2}~\cite{sentence_transformers_all_mpnet_base_v2}) and a simple 
magnitude based top-K selection strategy for SAE training. While sentence transformer 
embeddings provide moderate improvements 
over BERT and activation magnitude 
serves as a lightweight internal baseline, 
both remain consistently inferior 
to SAE-Embedding across models and domains.
(A broader comparison of Top-K coordinate selectors (random, magnitude, variance, PCA, and ours) is provided in the supplement; see Section~\ref{sec:sup_topk}.)

Using SAE embeddings consistently improves performance across models and domains. For Vicuna, the average gain after fine-tuning is +5.08 for the original model and +2.06 for the pruned model. This suggests that SAE-derived sparse, interpretable features better align with task-relevant representations, leading to higher-quality fine-tuning data.

\begin{table}[H]
\caption{Performance comparison of different dataset selection methods. Sentence-Transformer uses \texttt{all-mpnet-base-v2}. MC-CL: MathCoder-CL.}
\centering
\scriptsize
\setlength{\tabcolsep}{4pt}
\renewcommand{\arraystretch}{1}
\begin{tabular}{l l l c c c}
  \toprule
  Model & Type & Dataset Selection & STEM & Social Sci. & Hum. \\
  \midrule
  \multirow{6}{*}{Vicuna}  
    & Full   & BERT-Embedding & 31.17 & 35.04 & 34.49 \\
    & Full   & Sentence-Transformer & 32.08 & 36.72 & 36.15 \\
    & Full   & SAE-Embedding(Magnitude) & 30.92 & 34.10 & 33.85 \\
    & Full   & SAE-Embedding  & \textbf{33.32} & \textbf{40.21} & \textbf{42.43} \\
    & Pruned & BERT-Embedding & 29.01 & 34.23 & 34.48 \\
    & Pruned & Sentence-Transformer & 29.87 & 35.02 & 35.41 \\
    & Pruned & SAE-Embedding(Magnitude) & 28.74 & 33.51 & 33.02 \\
  \rowcolor{blue!13}
    & Pruned & SAE-Embedding  & \textbf{31.12} & \textbf{36.23} & \textbf{36.55} \\
  \midrule
  \multirow{6}{*}{MC-CL}  
    & Full   & BERT-Embedding & 30.45 & 29.14 & 29.01 \\
    & Full   & Sentence-Transformer & 31.82 & 30.02 & 29.88 \\
    & Full   & SAE-Embedding(Magnitude) & 29.73 & 27.95 & 27.60 \\
    & Full   & SAE-Embedding  & \textbf{34.96} & \textbf{32.91} & \textbf{31.66} \\
    & Pruned & BERT-Embedding & 24.91 & 12.73 & 12.53 \\
    & Pruned & Sentence-Transformer & 25.37 & 13.11 & 13.02 \\
    & Pruned & SAE-Embedding(Magnitude) & 23.84 & 11.92 & 11.74 \\
  \rowcolor{blue!13}
    & Pruned & SAE-Embedding  & \textbf{25.89} & \textbf{13.81} & \textbf{13.68} \\
  \midrule
  \multirow{6}{*}{LLaMa3.1}  
    & Full   & BERT-Embedding & 46.19 & 49.22 & 50.94 \\
    & Full   & Sentence-Transformer & 47.10 & 50.05 & 51.21 \\
    & Full   & Activation-Magnitude & 45.32 & 47.88 & 49.03 \\
    & Full   & SAE-Embedding  & \textbf{48.84} & \textbf{51.66} & \textbf{51.45} \\
    & Pruned & BERT-Embedding & 37.02 & 39.32 & 39.00 \\
    & Pruned & Sentence-Transformer & 38.11 & 40.08 & 40.14 \\
    & Pruned & Activation-Magnitude & 36.45 & 38.02 & 37.66 \\
  \rowcolor{blue!13}
    & Pruned & SAE-Embedding  & \textbf{39.91} & \textbf{41.07} & \textbf{41.19} \\
  \bottomrule
\end{tabular}
\label{tab:model_results_emb}
\end{table}

\section{Conclusion} 
We presented FineScope, a unified framework that enables efficient domain adaptation of LLMs by integrating sparse autoencoder-guided data selection with structured pruning and fine-tuning. By identifying semantically relevant samples from large unlabeled corpora, FineScope constructs compact, high-quality datasets that guide both compression and adaptation. Across diverse domains, FineScope consistently improves performance while significantly reducing model size, demonstrating that targeted data selection is crucial for effective LLM deployment in resource-constrained settings.

\onecolumn

\title{FineScope\\(Supplementary Material)}
\maketitle

This document provides additional methodological details, analyses, and ablations that support the claims of the main paper but are omitted there due to space constraints. In particular, we include (i) full objective and implementation details required for reproducibility, (ii) additional rationale for key design choices in our data-selection and pruning pipeline, and (iii) extended experimental results that probe robustness, generality, and sensitivity to hyperparameters.

The supplement is organized as follows. Section~\ref{sec:sup_sae_objective} specifies the complete sparse autoencoder (SAE) training objective and its components, complementing the SAE description in the main paper and enabling faithful replication.
In Section~\ref{sec:sup_rationale}, we discuss about our key design choices, why we use SAE codes as retrieval embeddings, select Top-$K$ coordinates via Jacobian sensitivity, and perform domain-conditioned structured pruning.
Section~\ref{sec:sup_breakdown} presents an ablation over different Top-$K$ selection strategies for SAE training, supporting the claim that Jacobian-based selection is an important ingredient rather than other heuristics.
Section~\ref{sec:sup_pruning} evaluates multiple pruning methods under matched pruning ratios, supporting the claim that our improvements are data-centric and persist across pruning algorithms.
Section~\ref{sec:sup_qwen} reports results on additional model families (Qwen variants) to examine how FineScope generalizes beyond a single backbone.
Section~\ref{sec:sup_compbreak} provides a computation-cost breakdown of the Top-$K$ scoring and SAE training stages, discussing the practical overhead of the proposed pipeline.
Section~\ref{sec:sup_pruning_ratio} studies how FineScope behaves over various pruning ratios, showing that FineScope maintains performance under more aggressive pruning than baseline data selection
Sections~\ref{sec:sup_seed} analyzes sensitivity to the number of initial seeds and the number of selected seeds, respectively, showing that FineScope is effective with small user input and clarifying stable operating regimes.
Finally, Sections~\ref{sec:sup_seed_qual} provides qualitative analysis and concrete seed examples, illustrating the semantics captured by our representations.

\section{Sparse Autoencoder Objective Details}
\label{sec:sup_sae_objective}

This subsection describes the training setup for the SAE discussed in main Section~\ref{sae_training} to make it fully reproducible with detailed definitions of the activation tensor, SAE encoder/decoder parameterization, and residual used in the loss.

\subsection{Setup}
\label{sec:sup_setup}
Let $a \in \mathbb{R}^{d}$ denote an activation vector extracted from a fixed (frozen) pretrained transformer at a chosen hookpoint in layer $\ell$.
For a mini-batch $A \in \mathbb{R}^{B \times d}$, the sparse autoencoder (SAE)\footnote{Loss implementations are from~\cite{EleutherAI_SAE} repository.} produces a reconstruction
\begin{equation}
\hat{A} = \mathrm{SAE}(A),
\end{equation}
and we define the residual
\begin{equation}
E = \hat{A} - A.
\end{equation}
The encoder produces latent activations
\begin{equation}
Z = \mathrm{Enc}(A) = \mathrm{ReLU}(W_e A^\top + b_e)^\top \in \mathbb{R}^{B \times d_{\text{sae}}}.
\end{equation}

\subsection{Scale-Normalized Reconstruction Loss (FVU)}
\label{sec:sup_fvu}
We optimize a scale-normalized reconstruction objective based on the fraction of variance unexplained (FVU).
We use an FVU-style reconstruction objective to make reconstruction errors comparable across layers and activation scales, which stabilizes SAE training when applying the method to multiple layers as discussed in main Section~\ref{sae_training}.

Let $\mu \in \mathbb{R}^{d}$ be the feature-wise batch mean:
\begin{equation}
\mu = \frac{1}{B}\sum_{i=1}^{B} A_i.
\end{equation}
Define the total variance term
\begin{equation}
V = \left\lVert A - \mathbf{1}\mu^\top \right\rVert_F^2,
\end{equation}
where $\mathbf{1} \in \mathbb{R}^{B}$ is the all-ones vector and $\|\cdot\|_F$ is the Frobenius norm.
The FVU loss is
\begin{equation}
L_{\mathrm{FVU}} = \frac{\|\hat{A} - A\|_F^2}{\|A - \mathbf{1}\mu^\top\|_F^2} = \frac{\|E\|_F^2}{V}.
\label{eq:fvu}
\end{equation}

This normalization reduces sensitivity to the absolute activation magnitude and makes the sparsity/auxiliary-loss weights more transferable across layers.

\subsection{Auxiliary Dead-Latent Residual Loss (AuxK)}
\label{sec:sup_auxk}
To ensure the SAE code provides a usable embedding for retrieval (main Section~\ref{datacurate}), we mitigate `dead' latents, i.e., features rarely activated, which would otherwise contribute little to representation capacity.
To this end, we add an auxiliary loss that encourages dead latents to predict the residual.
Let $m \in \{0,1\}^{d_{\text{sae}}}$ be a dead-latent mask, and define
\begin{equation}
n_{\text{dead}} = \sum_{j=1}^{d_{\text{sae}}} m_j.
\end{equation}
When $n_{\text{dead}}>0$, we select up to $k_{\text{aux}}$ dead latents and decode only those to obtain an auxiliary residual prediction $\hat{E}$.
We use the scaling factor
\begin{equation}
s = \min\left(\frac{n_{\text{dead}}}{k_{\text{aux}}}, 1\right),
\end{equation}
and define
\begin{equation}
L_{\mathrm{AuxK}} = s \cdot \frac{\|\hat{E} - E\|_F^2}{V}.
\label{eq:auxk}
\end{equation}
If no dead latents are present, we set $L_{\mathrm{AuxK}}=0$.

Empirically, this stabilizer improves feature utilization so the embedding space does not collapse onto a small subset of active units.

\subsection{Sparsity Regularization}
\label{sec:sup_l1}

To explicitly enforce sparsity in the SAE latents, we add an $\ell_1$ penalty on the (post-ReLU) encoder activations.
Using the $\ell_1$ penalty improves interpretability and makes cosine similarity in code space reflect overlap in a small set of salient features for the embedding argument discussed in main Section~\ref{datacurate}, while also reducing the chance that similarity is dominated by diffuse, low-signal dimensions:

\begin{equation}
L_{\ell_1} = \|Z\|_1 = \sum_{i=1}^{B}\sum_{j=1}^{d_{\text{sae}}} |Z_{i,j}|.
\label{eq:l1}
\end{equation}

\subsection{Final SAE Objective}
\label{sec:sup_final_objective}
Combining reconstruction, dead-latent auxiliary loss, and explicit sparsity regularization, the final SAE objective is:
\begin{equation}
L_{\mathrm{SAE}} = L_{\mathrm{FVU}} + \alpha L_{\mathrm{AuxK}} + \lambda L_{\ell_1},
\label{eq:sae_final}
\end{equation}
where $\alpha \ge 0$ and $\lambda \ge 0$ are hyperparameters.

\section{Design Rationale: Embeddings, Top-K Selection, and Pruning} \label{sec:sup_rationale}

This section provides theoretical perspective on three design choices in FineScope: 
(i) using SAE encoder codes as embeddings for domain-aligned retrieval (Section~\ref{Method}), 
(ii) using Jacobian sensitivity to select a compact set of activation coordinates for SAE training (Section~\ref{sae_training}), and 
(iii) conditioning structured pruning on the curated domain dataset $D_s$ (Section~\ref{datacurate}). 

\subsection{SAE Encoder Codes as Domain-Aligned Embeddings} \label{sec:sup_sae}
We first instantiate the embedding used in main Section~\ref{datacurate} and clarifies exactly what vector is used for similarity-based retrieval.

\paragraph{Embedding definition}
In Section~\ref{datacurate}, we denote embeddings with $\mathrm{SAE}(x)$ when computing similarity (Eq~\ref{eq:similarity}).
Concretely, we use the SAE \emph{encoder code} as the embedding:
\[
z(x) \;=\; \mathrm{Enc}(\tilde{a}^{(\ell)}(x)),
\]
\noindent where $\tilde{a}^{(\ell)}(x)$ is the pooled activation vector restricted to the selected coordinate set $I^{(\ell)}$ (Eq~\ref{eq:filtered_activation_main}).
Cosine similarity is computed in code space, i.e., $\mathrm{CosSim}(z(x_u), z(x_t))$.
This choice supports the main claim that FineScope retrieves data using a representation aligned with the internal features rather than surface-form similarity.

\paragraph{Sparse-coding viewpoint.}
We here provide a sparse-coding interpretation to explain why distances in code space can capture semantically meaningful feature overlap, which is the mechanism FineScope uses for domain-aligned retrieval.
An SAE trained with reconstruction plus an $\ell_1$ penalty (Eq~\ref{eq:loss}) can be viewed as a form of sparse dictionary learning: the decoder columns define a feature dictionary, and the encoder produces sparse coefficients selecting a small number of those features.
Under a standard generative approximation, the intermediate activations admit a decomposition
\[
a^{(\ell)}(x) \;\approx\; D z^\star(x) + \epsilon,
\]
\noindent where $D$ is a (possibly overcomplete) dictionary, $z^\star(x)$ is sparse, and $\epsilon$ is residual noise.
Training the SAE approximately recovers a coordinate system in which the signal is concentrated in a small number of latent features.
Consequently, under this view, nearest-neighbor retrieval selects examples sharing latent feature activations that the model already uses internally.

\paragraph{Domain alignment perspective.}
Assume there exists a latent subset of features $S_{\mathrm{dom}}$ that are preferentially active for a target domain distribution $P_{\mathrm{dom}}$ and less active under a background distribution $P_{\mathrm{bg}}$.
Then for two samples $x, x'$ from the same domain, their sparse codes share support more often:
\[
\mathbb{P}[\mathrm{supp}(z^\star(x)) \cap \mathrm{supp}(z^\star(x')) \neq \emptyset \mid x,x'\sim P_{\mathrm{dom}}]
\;>\;
\mathbb{P}[\cdot \mid x\sim P_{\mathrm{dom}}, x'\sim P_{\mathrm{bg}}].
\]
Cosine similarity in sparse code space increases with (i) overlap in active features and (ii) alignment in their coefficient patterns.
Thus, nearest-neighbor retrieval in $z(x)$ space preferentially selects samples that activate similar internal features which is a representation-level notion of semantic similarity tied to the pretrained model's internal organization rather than surface-form overlap.
This supports that the retrieved dataset $D_s$ concentrates \textit{domain-relevant behavior} by selecting samples that activate the same internal feature subset as the seeds.

\paragraph{Relation to seed-based retrieval in Eq.~\ref{eq:similarity}}
Eq.~\ref{eq:similarity} uses $s(x_u)=\max_{x_t\in D_t}\mathrm{CosSim}(z(x_u),z(x_t))$, which can be interpreted as a one-vs-set score:
it selects $x_u$ if it has high feature overlap with \emph{at least one} target seed.
This is well-matched to small-seed settings where the target domain may be multimodal (multiple subtopics); the max operator acts like a mixture-of-modes retrieval rule.
This interpretation directly supports our small-seed setting claim: the max-over-seeds scoring behaves like a multi-prototype retriever when the target domain is heterogeneous

\subsection{Jacobian Sensitivity as a Coordinate-Selection Criterion} \label{sec:sup_jacob}
\paragraph{Problem setting.}
Training an SAE on the full activation width $d_\ell$ is expensive, so we choose a subset of coordinates $I^{(\ell)}$ of size $K_{\mathrm{act}}\ll d_\ell$ (Section~\ref{topk}).
It determines which parts of the representation the SAE can model under a fixed budget, and thus directly affects downstream retrieval quality (as we further evaluate in Section~\ref{sec:sup_breakdown})

The coordinate score is based on input sensitivity (Eq.~\ref{eq:saliency_per_example}):
\[
s^{(\ell)}_j(x) = \left\|\frac{\partial a^{(\ell)}_j(x)}{\partial e(x)}\right\|_F.
\]

\paragraph{Estimator property.}
Let $J^{(\ell)}(x)=\partial a^{(\ell)}(x)/\partial \mathrm{vec}(e(x))$ and sample Rademacher probes $r\in\{\pm1\}^{Td_{\mathrm{model}}}$ with $\mathbb{E}[rr^\top]=I$.
For a JVP $u=J^{(\ell)}(x)r$ (Eq.~\ref{eq:jvp_def}), the per-coordinate second moment satisfies:
\[
\mathbb{E}[u_j^2]
= \mathbb{E}[(J^{(\ell)}_{j,:}r)^2]
= J^{(\ell)}_{j,:}\,\mathbb{E}[rr^\top]\,(J^{(\ell)}_{j,:})^\top
= \|J^{(\ell)}_{j,:}\|_2^2.
\]
Therefore the estimator in Eq.~(8), $\hat{s}_j^2=\frac{1}{R}\sum_{p=1}^R (u_j^{(p)})^2$, is unbiased for the squared Jacobian row norm.
Independence of probes implies the estimation variance decreases on the order of $1/R$.
This makes Jacobian-based scoring practical as we can estimate sensitivity using JVPs without explicitly forming the Jacobian, enabling scalable Top-$K$ selection in main Section~\ref{topk}.

\paragraph{Jacobian row norms are a principled selection signal.}
A coordinate with small $\|J^{(\ell)}_{j,:}(x)\|_2$ is locally insensitive to perturbations in the input embedding: it varies weakly with input content.
Such coordinates contribute little to distinguishing inputs in a local linearization and are less useful for building a compact representation space for retrieval.
Conversely, large row norm indicates that coordinate $j$ is \emph{functionally responsive} to input variation, suggesting it carries more information about prompt-dependent semantics.

A useful lens is a local linear approximation around $e(x)$:
\[
a^{(\ell)}(e+\delta) \approx a^{(\ell)}(e) + J^{(\ell)}(x)\,\mathrm{vec}(\delta).
\]
If we retain only a coordinate subset $I$, the best linear reconstruction of the full activation variation from those coordinates is limited by how much of $J^{(\ell)}(x)$ lies in the rows indexed by $I$.
Selecting coordinates with large row norms is a greedy strategy for retaining directions with large local sensitivity, i.e., a larger share of the input-induced activation variation.

Intuitively, selecting large-row-norm coordinates preserves more of the input-conditioned variation that the embedding must reflect for similarity search to work.
This directly aligns with the goal of learning an SAE embedding that meaningfully varies across inputs (needed for similarity-based retrieval).

\paragraph{Why this criterion is preferable to purely activation-based heuristics.}
Magnitude/variance heuristics select coordinates that are large or highly varying under a reference distribution, but they do not distinguish between variation induced by input semantics versus variation induced by internal noise, positional effects, or distributional artifacts.
Jacobian sensitivity explicitly measures \emph{causal responsiveness} of the coordinate to the input embedding, which is closer to the notion of semantic dependence needed for representation-based retrieval.
We validate this motivation empirically by comparing against magnitude/variance and PCA-based alternatives in later Section~\ref{sec:sup_breakdown}.

\subsection{Domain-Conditioned Structured Pruning and Domain-Critical Components}\label{sec:sup_domain}
\paragraph{Pruning objective viewpoint.}
This subsection justifies the main paper’s design choice to compute pruning importance on the curated domain dataset $D_s$ (Section~\ref{pruningtuning}), rather than on a generic corpus.
Structured pruning can be framed as selecting a restricted architecture that minimizes loss on a target distribution.
Let $P_{\mathrm{dom}}$ denote the target domain distribution, and let $\mathcal{L}(M; x)$ denote the per-sample loss of model $M$.
An ideal domain-specific pruning objective is:
\[
\min_{M'\in\mathcal{M}(r)} \; \mathbb{E}_{x\sim P_{\mathrm{dom}}}\big[\mathcal{L}(M'; x)\big],
\]
where $\mathcal{M}(r)$ denotes models obtainable by pruning at ratio $r$ (Eq.~\ref{eq:pruning}).
In practice $P_{\mathrm{dom}}$ is unknown and $D_s$ is used as a proxy sample from it.

\paragraph{First-order approximation behind gradient-based importance.}
Many structured pruning methods score components (blocks/heads/MLPs) by estimating the loss change if a component is removed or down-scaled.
Consider a block output $h_b$ and an (implicit) gating parameter $\alpha_b$ that scales the block (pruning corresponds to $\alpha_b\to 0$).
Then by Taylor expansion:
\[
\Delta \mathcal{L}_{D_s} \approx 
\left.\frac{\partial \mathcal{L}_{D_s}}{\partial \alpha_b}\right|_{\alpha_b=1}\,(\alpha_b-1),
\quad\text{with}\quad
\frac{\partial \mathcal{L}_{D_s}}{\partial \alpha_b}
= \left\langle \frac{\partial \mathcal{L}_{D_s}}{\partial h_b},\, h_b \right\rangle.
\]
This motivates block importance scores that aggregate gradient-based attributions over $D_s$.
Under this approximation, removing blocks with small importance has small predicted effect on the domain loss, while retaining large-importance blocks preserves domain-relevant computation.

\paragraph{Why conditioning on $D_s$ changes what is preserved.}
If importance is computed on a generic distribution $P_{\mathrm{gen}}$, the resulting retained components favor generic behaviors.
Conditioning importance on $D_s$ (intended to approximate $P_{\mathrm{dom}}$) shifts the scoring toward components whose outputs most influence domain loss.
Thus the retained structure can be interpreted as \emph{domain-critical components under ratio $r$}, i.e., those that matter most for the domain objective among candidates removable by the pruning scheme.
This supports the main experimental finding that domain-conditioned pruning improves post-pruning domain accuracy relative to pruning guided by generic tuning data.

\paragraph{Consistency under distribution approximation.}
Let $g_b(x)$ denote the per-sample importance contribution for block $b$ (e.g., a gradient-attribution score).
If $D_s$ is an i.i.d. sample from $P_{\mathrm{dom}}$, then by standard concentration, the empirical average $\frac{1}{|D_s|}\sum_{x\in D_s} g_b(x)$ concentrates around $\mathbb{E}_{x\sim P_{\mathrm{dom}}}[g_b(x)]$.
Therefore, scoring blocks on $D_s$ approximates scoring them on the true domain distribution, and pruning becomes a principled approximation to domain-specific architecture selection.

\paragraph{Terminology note.}
The term ``domain subnetwork'' can be understood as the set of components retained by domain-conditioned importance estimation under a fixed pruning ratio.

We do not assume uniqueness of this set; different pruning procedures or tie-breaking can yield different retained sets with similar loss.
For clarity, we refer to these as \emph{domain-critical components} preserved by the pruning rule at ratio $r$.

\section{Effect of Different Top-K selection method for SAE training} \label{sec:sup_breakdown} \label{sec:sup_topk}

Table~\ref{tab:coordinate_selectors} studies how the \emph{coordinate selector} used in the Top-$K$ activation preselection step affects downstream performance when training SAEs.
This directly supports main Section~\ref{topk} by isolating the effect of the coordinate selector while holding the SAE architecture and training budget fixed.
Because training on the full activation width $d_\ell$ is expensive, we preselect a fixed subset of $K_{\text{act}} \ll d_\ell$ activation coordinates per layer and train the SAE only on those coordinates. 

We compare selectors that represent common, progressively more informed choices: (i) \textbf{Random} uniform selection as a sanity-check lower bound, (ii) \textbf{Activation Magnitude} and \textbf{Activation Variance} as lightweight activation-based heuristics computed over a reference corpus $D_0$, and (iii) a stronger unsupervised baseline based on \textbf{PCA subspace projection} (top-$K_{\text{act}}$ principal components), which captures global covariance structure rather than making independent coordinate-wise decisions. 
These baselines accordingly cover uninformed selection, common activation-statistic heuristics, and global subspace methods, which are the most plausible alternatives.

Performance improves monotonically with selector sophistication: random selection performs worst, activation statistics provide consistent gains, PCA subspace projection improves further, and our method achieves the best results across all three subdomains (12.91/10.12/7.01 on Pre-algebra/Algebra/C\&P). Notably, the largest gaps appear in \emph{Algebra} and \emph{Counting and Probability}, suggesting that choosing semantically informative coordinates is particularly important for preserving reasoning-relevant structure under aggressive dimensionality reduction.

Our selector ranks coordinates by \emph{input sensitivity}, based on how strongly each activation coordinate responds to perturbations in the (frozen) input embedding activations.
This criterion targets coordinates that are not merely large or variable, but \emph{causally responsive} to input-dependent semantics: precisely the signal needed for learning an SAE embedding space that supports downstream retrieval and domain alignment. In contrast, magnitude/variance heuristics can over-select coordinates whose variability is driven by scale, noise, or distributional artifacts rather than meaningful input dependence. 

Overall, the results indicate that incorporating first-order sensitivity information during Top-$K$ selection yields a consistently stronger representation for SAE training than purely activation-driven or unsupervised subspace alternatives.
 
\begin{table}[h!]
\centering
\caption{Coordinate selector comparison at fixed $K_{\text{act}}$. 
All methods use the same SAE architecture and training budget. 
For PCA, we explicitly compare subspace projection (top-$K_{\text{act}}$ principal components) rather than coordinate masking.}
\label{tab:coordinate_selectors}
\begin{tabular}{lccc}
\toprule
Selector & Pre-algebra & Algebra & Counting and Probability \\
\midrule
Random (uniform)                  & 7.12 & 4.88 & 3.41 \\
Activation Magnitude ($\mathbb{E}_{x\sim D_0}[|a_j|]$) & 8.94 & 6.22 & 4.37 \\
Activation Variance ($\mathrm{Var}_{x\sim D_0}(a_j)$)  & 9.31 & 6.58 & 4.64 \\
PCA Subspace (Top-$K_{\text{act}}$ PCs)                & 10.05 & 7.42 & 5.11 \\
\midrule
Ours & 12.91 & 10.12 & 7.01 \\
\bottomrule
\end{tabular}
\end{table}

\section{Different Pruning Methods}\label{sec:sup_pruning}
Table~\ref{tab:pruning_baselines} compares three representative pruning baselines: Magnitude Pruning~\cite{wanda}, FLAP~\cite{flap}, and LLMPruner~\cite{llmpruner} under a controlled setting where the pruning ratio is matched and only the \emph{pruning/tuning data} is varied (Alpaca vs.\ FineScope).
We include these strategies to span a spectrum of widely used pruning behaviors: a simple magnitude-based baseline that is data-agnostic and commonly used as a first reference point, alongside stronger LLM-oriented structured pruning baselines (FLAP, LLMPruner) that are designed to better preserve functional components under compression. This selection lets us test whether FineScope’s benefit is specific to a particular pruning rule or instead transfers across qualitatively different pruning objectives.

Consistent with our domain-conditioned pruning rationale shown in Section~\ref{sec:sup_domain}, using $D_s$ shifts importance estimation toward domain-critical components.
From an optimization perspective, domain-conditioned pruning can be viewed as approximating the minimization of expected loss under the target distribution, with $D_s$ serving as a proxy sample; consequently, importance estimates shift toward \emph{domain-critical} components that would be undervalued under a generic pruning distribution. The sharp Alpaca $\rightarrow$ FineScope improvements in Algebra and Counting further suggest that generic pruning data disproportionately removes circuitry needed for compositional/algorithmic reasoning, whereas domain-aligned pruning data better preserves these specialized computations during compression. 

\begin{table}[h!]
\centering
\caption{Comparison of different pruning methods with different tuning/pruning dataset}
\label{tab:pruning_baselines}
\begin{tabular}{lccc}
\toprule
Method  & Pre-algebra & Algebra & Counting and Probability \\
\midrule
\multicolumn{4}{c}{\textbf{Alpaca}} \\
\midrule
Magnitude Pruning                  & 3.89 & 0.25 & 0.18 \\
FLAP                          & 7.02 & 0.45 & 0.34 \\
LLMPruner                           & 5.56 & 0.30 & 0.21 \\
\midrule
\multicolumn{4}{c}{\textbf{FineScope}} \\
\midrule
Magnitude Pruning                  & 8.74 & 6.12 & 4.01 \\
FLAP                          & 12.15 & 9.09 & 6.47 \\
\textbf{LLMPruner}                           & \textbf{12.91} & \textbf{10.12} & \textbf{7.01} \\
\bottomrule
\end{tabular}
\end{table}

\section{Experiments on a Different Model Family: Qwen} \label{sec:sup_qwen}
To assess generality beyond the model family used in the main experiments, we apply FineScope to multiple Qwen generations under the same pipeline.
In Table~\ref{tab:qwen}, we implemented FineScope algorithm on three different Qwen models: Qwen 3-8B~\cite{qwen3}, Qwen 2.5-7B~\cite{qwen25} and Qwen 2-7B~\cite{qwen2}. FineScope consistently improves performance over the Alpaca baseline across all evaluated Qwen models and mathematical domains. This trend suggests that the FineScope pipeline remains beneficial even when the overall accuracy is relatively low, indicating that the gains are not limited to only high-capacity or high-performing settings. Overall, these results suggest that FineScope provides stable and consistent benefits across different generations of Qwen models. Another notable pattern is that the relative gains are larger for the older and weaker models than for Qwen 3-8B, implying that targeted data curation may be especially valuable when the base model has more limited inherent reasoning ability.

\begin{table}[h!]
\centering
\caption{Performance analysis of FineScope pipeline on Qwen models}
\label{tab:qwen}
\begin{tabular}{lccc}
\toprule
Models  & Pre-algebra & Algebra & Counting and Probability \\
\midrule
\multicolumn{4}{c}{\textbf{Alpaca}} \\
\midrule
Qwen 3-8B                  & 15.00 & 12.31 & 11.01 \\
Qwen 2.5-7B                          & 9.91 & 8.30 & 7.31 \\
Qwen 2-7B                         & 9.50 & 4.40 & 6.10\\
\midrule
\multicolumn{4}{c}{\textbf{FineScope}} \\
\midrule
Qwen 3-8B                  & 17.31 & 14.22 & 12.80\\
Qwen 2.5-7B                          & 15.21 & 11.24 & 12.10 \\
Qwen 2-7B                         & 11.89 & 12.10 & 10.32\\
\bottomrule
\end{tabular}
\end{table}

\section{Computation cost} \label{sec:sup_compbreak}
\begin{figure*}[h!]
    \centering
    \includegraphics[width=1\linewidth]{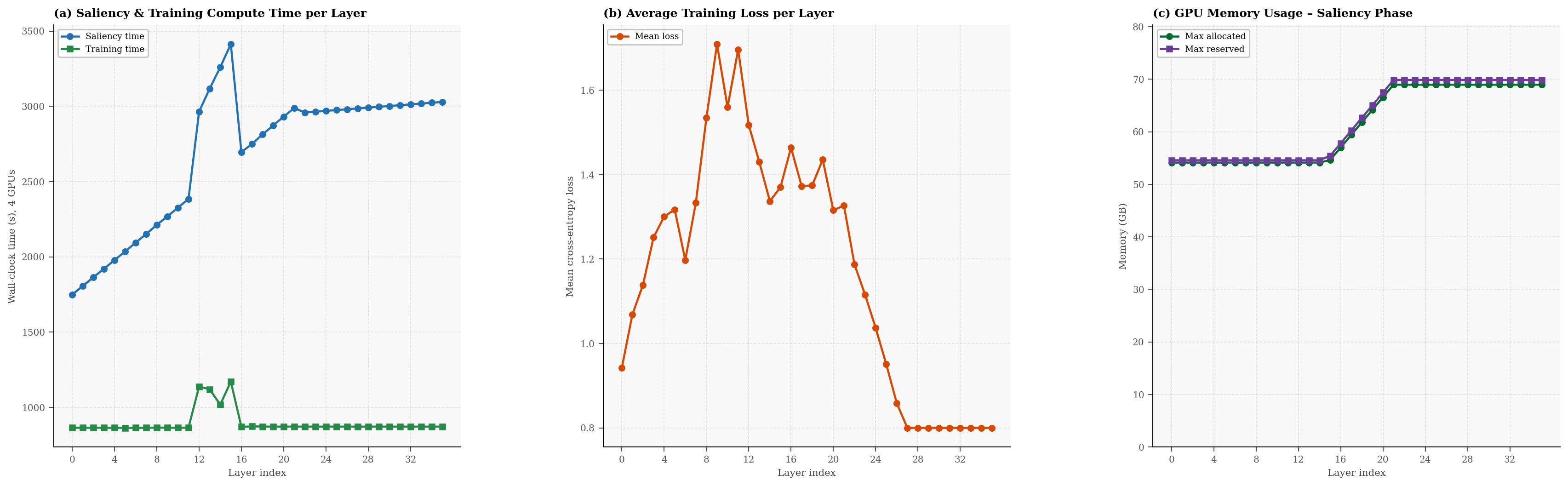}
    \caption{Computational cost of Qwen 8B through-out all layers}
    \label{fig:selected_profile}
\end{figure*}

This section discusses the practical overhead of FineScope by breaking down runtime across layers, supporting the claim that the method is feasible under realistic budgets.
Figure~\ref{fig:selected_profile} reports a per-layer runtime breakdown for Qwen 8B on four GPUs, separating the \emph{saliency} stage (JVP-based Top-$K$ coordinate scoring) from \emph{SAE training}.
A key observation is that saliency computation dominates wall-clock time across layers, consistently exceeding the SAE optimization cost. Moreover, saliency time generally increases with depth, indicating that the overall runtime is governed more by the cost of producing the saliency signal than by the SAE update itself.

The transient nature of the peak and its concentration in the saliency component indicate a \emph{systems-level transition} that predominantly impacts JVP execution.
Such behavior arises from a shift in kernel selection or autotuning decisions, a temporary fallback to a suboptimal execution path requiring additional tensor materialization, or allocator-induced effects that introduce synchronization overhead.
Since the spike is not sustained, it is most consistent with one-off initialization or execution-path selection costs rather than a structural increase in per-layer arithmetic.

Overall, the profile indicates that even with a constant Top-$K$ size, end-to-end computational cost for Qwen 8B is primarily determined by the saliency/JVP pipeline and layer-dependent execution overhead, while SAE training contributes a smaller and comparatively more uniform fraction of runtime.

\section{Extended Pruning Ratio}   \label{sec:sup_pruning_ratio}
To understand how FineScope works over different pruning ratio and examine the limits of the model compression, we compared performance under different pruning ratios.
The experiments were conducted with SelfInstruct dataset on the LLaMA 3.1 8B model.

\begin{figure*}[h!]
    \centering
    \includegraphics[width=0.4\linewidth]{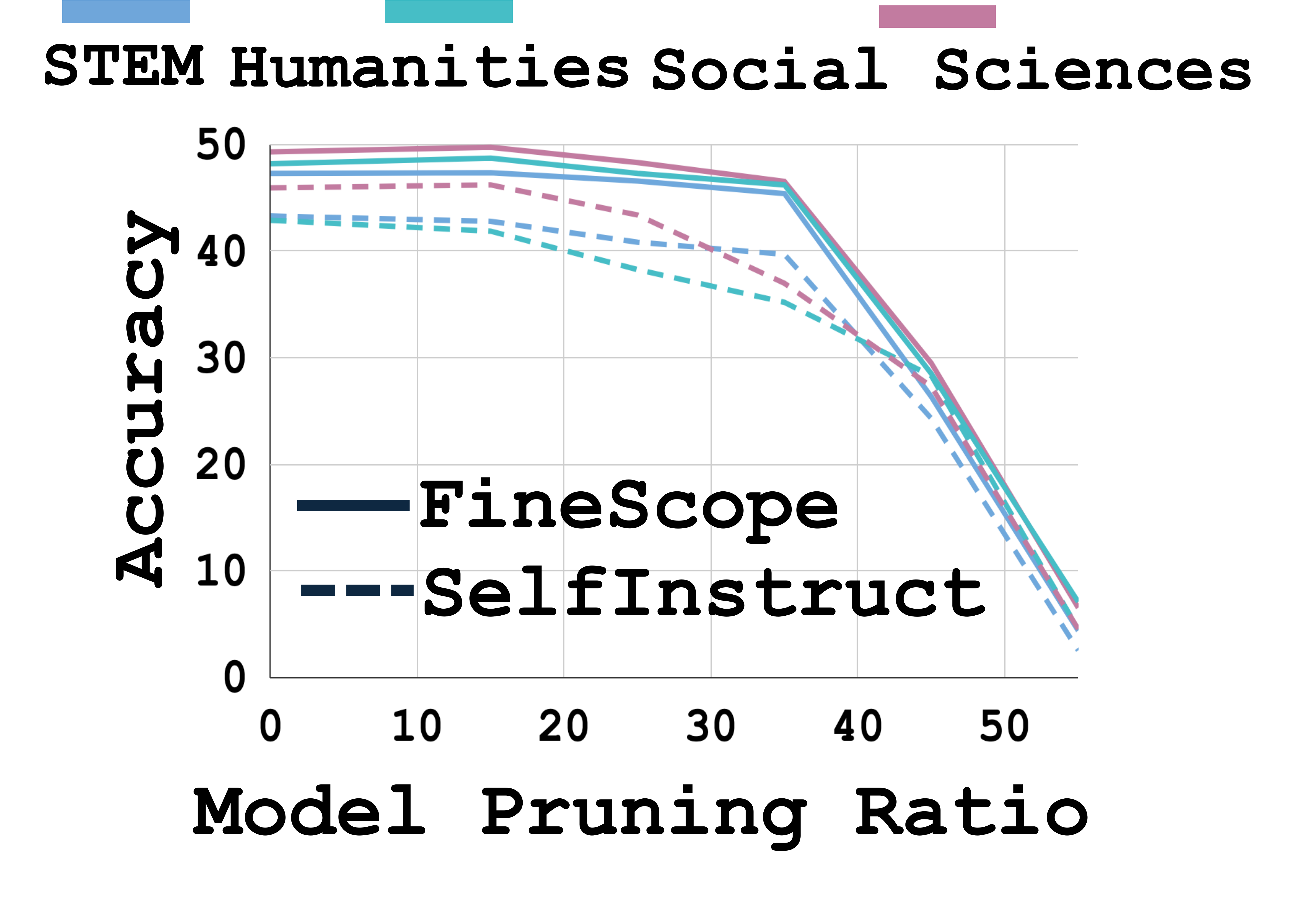}
    \caption{Effect of model pruning ratio on accuracy.}
    \label{fig:prunignratio}
\end{figure*}

As shown in Figure~\ref{fig:prunignratio}, FineScope demonstrates greater resilience to aggressive pruning. Although both methods show a decline in accuracy as pruning increases, FineScope consistently outperforms SelfInstruct across all settings. Notably, SelfInstruct begins to degrade at just 25\% pruning, while FineScope maintains stable accuracy up to 35\%. This indicates that FineScope enables higher pruning tolerance without compromising task performance.
This supports the claim that FineScope enables more aggressive pruning before accuracy collapses, compared to a larger but less targeted baseline dataset.

\section{Discussion for Seed Choice Sensitivity} \label{sec:sup_seed}
\subsection{Effect of Number of Initial Seeds}

\begin{figure}[h]
  \centering
  \includegraphics[width=0.5\textwidth]{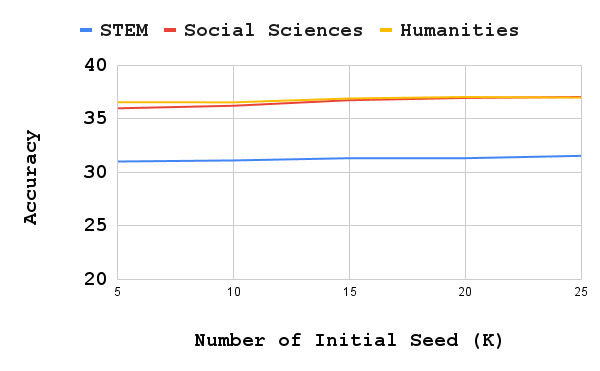}
  \caption{Effect of number of initial seeds on downstream task accuracy}
  \label{fig:initial_seeds}
\end{figure}

To verify how FineScope works in small-seed settings, we conduct a sensitivity analysis by varying the number of user-provided initial seeds.
Figure~\ref{fig:initial_seeds} shows that changing the number of initial seeds $K$ from 5 to 25 has only a marginal effect on accuracy across all domains. STEM accuracy remains nearly constant (around $\sim31\%$) with only a slight upward trend as $K$ increases. Social Sciences and Humanities consistently achieve higher accuracy (around $\sim36$-$37\%$), with a small improvement up to approximately $K=15$-$20$, after which performance largely plateaus.

Overall, the results indicate that accuracy is not very sensitive to the choice of $K$ in this range, suggesting that a modest number of initial seeds is sufficient.
This suggests that our proposed FineScope does not require heavy user input to produce effective domain-curated data.

\subsection{Effect of Number of Selected Seeds}
Next, we analyze how the number of selected seeds influence on the final performance. 
Here, `selected seeds' refers to the expanded set chosen by the pipeline after retrieval/selection, which controls how many prototypes guide subsequent data collection.

\begin{figure}[h]
  \centering
  \includegraphics[width=0.5\textwidth]{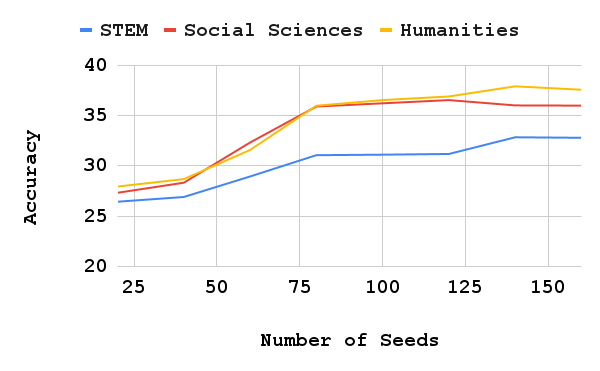}
  \caption{Effect of number of selected seeds on downstream task accuracy}
  \label{fig:selected_seeds}
\end{figure}

As shown in Figure~\ref{fig:selected_seeds}, increasing the number of seeds from 25 to around 80 yields a substantial accuracy gain across all domains, with the steepest improvements occurring between roughly 40 and 80 seeds (particularly for Social Sciences and Humanities). Beyond $\sim80$ seeds, gains begin to saturate: STEM improves gradually and then stabilizes, Social Sciences peaks around $\sim100$-$125$ seeds and slightly declines at higher values, while Humanities continues to improve up to about $\sim140$ seeds before showing a marginal drop.

This suggests diminishing returns after $\sim80$ selected seeds, where adding more seeds provides only limited (and sometimes unstable) improvements. Thus, we exploit the moderate selection size in FineScope considering the fixed computation budget.

\section{Qualitative Analysis for Seed Choice} \label{sec:sup_seed_qual}
\subsection{Seed Selection}  
\begin{figure*}[!h]
    \centering
    \includegraphics[width=1\linewidth]{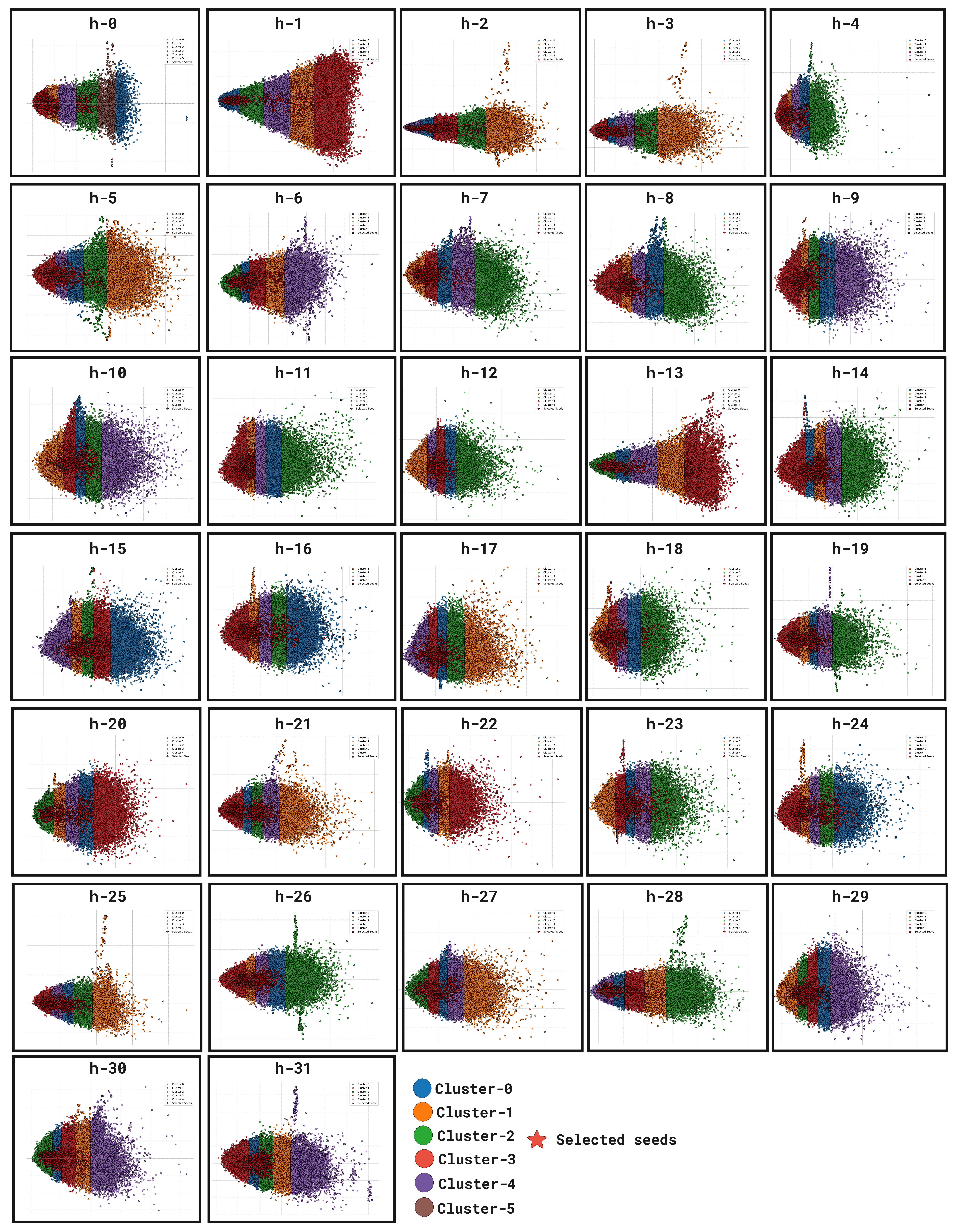}
    \caption{Cluster visualization for seed selection for dataset curation (for STEM)}
    \label{fig:seedselect_extend}
\end{figure*}

In this section, we include a qualitative analysis to make the seed selection mechanism more transparent.
We visualize the seed selection process using cluster analysis, as shown in \textbf{Figure~\ref{fig:seedselect_extend}}. The interpretation of features varies across initial, middle, and final layers, reflecting how representations evolve within the model. The results suggest that our curated dataset is alinged with the clustered virtual domain target domains across different layer representations. Lower layers tend to capture broad, generalized features, often encoding syntactic structures or common linguistic patterns. In contrast, deeper layers focus on increasingly abstract and domain-specific attributes, leading to more compact and semantically meaningful clusters. 

This progressive refinement suggests that domain-relevant information emerges more distinctly in later layers, where feature representations become more specialized. Our method leverages this hierarchical structure, selecting seeds from layers that balance generalization and domain specificity. By utilizing SAE-based representations instead of raw embeddings, we ensure that the seed selection process is more interpretable and aligned with high-level domain knowledge, rather than being influenced by superficial token-level similarities.
This suggests that SAE-guided representations yield more semantically aligned and thus more reliable retrieval than raw embedding similarity.

\subsection{Example Seeds}
\begin{figure*}
    \centering
    \includegraphics[width=1\linewidth]{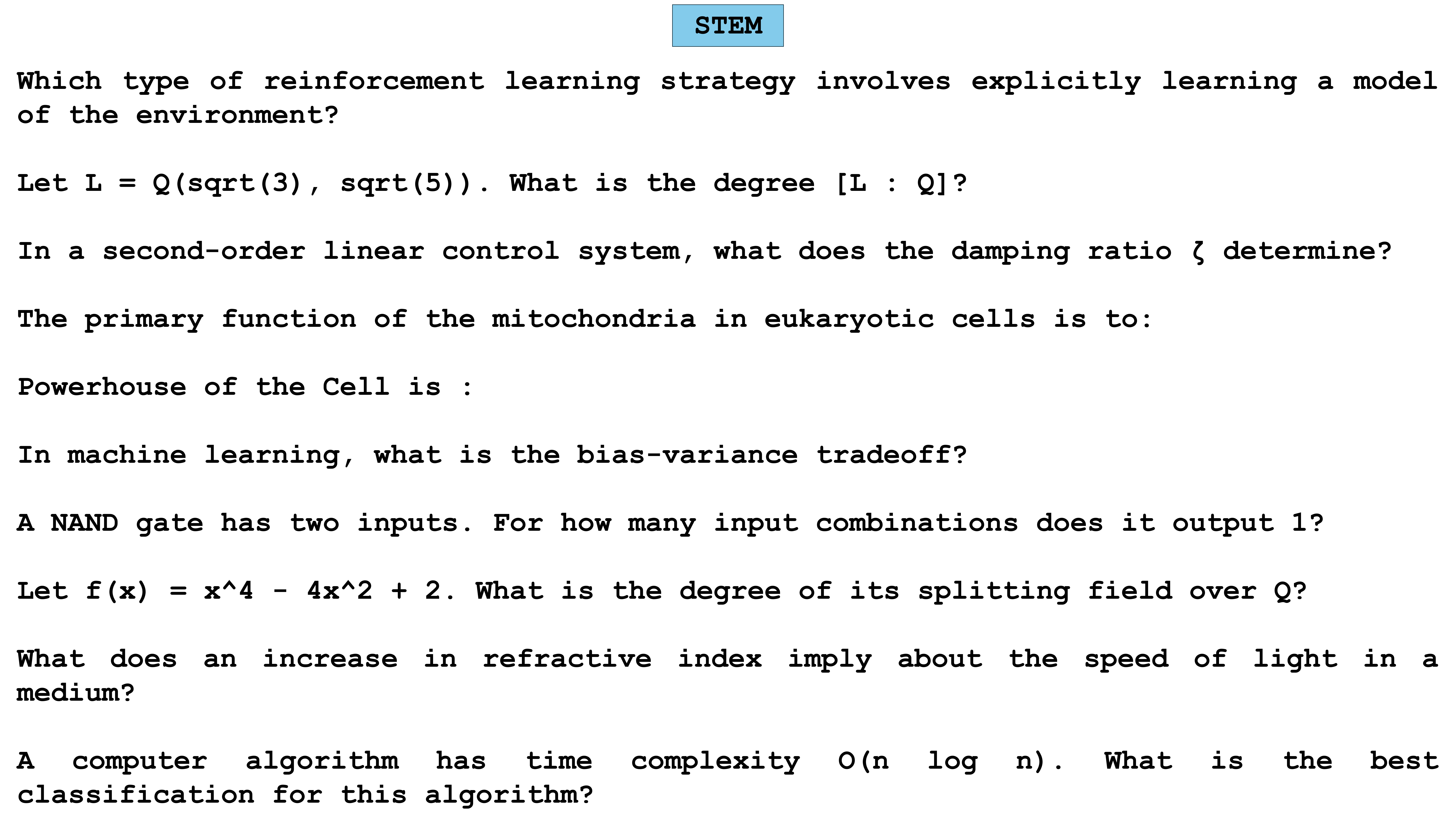}
    \caption{User defined seed sample of STEM target domain}
    \label{fig:stem}
\end{figure*}

\begin{figure*}
    \centering
    \includegraphics[width=1\linewidth]{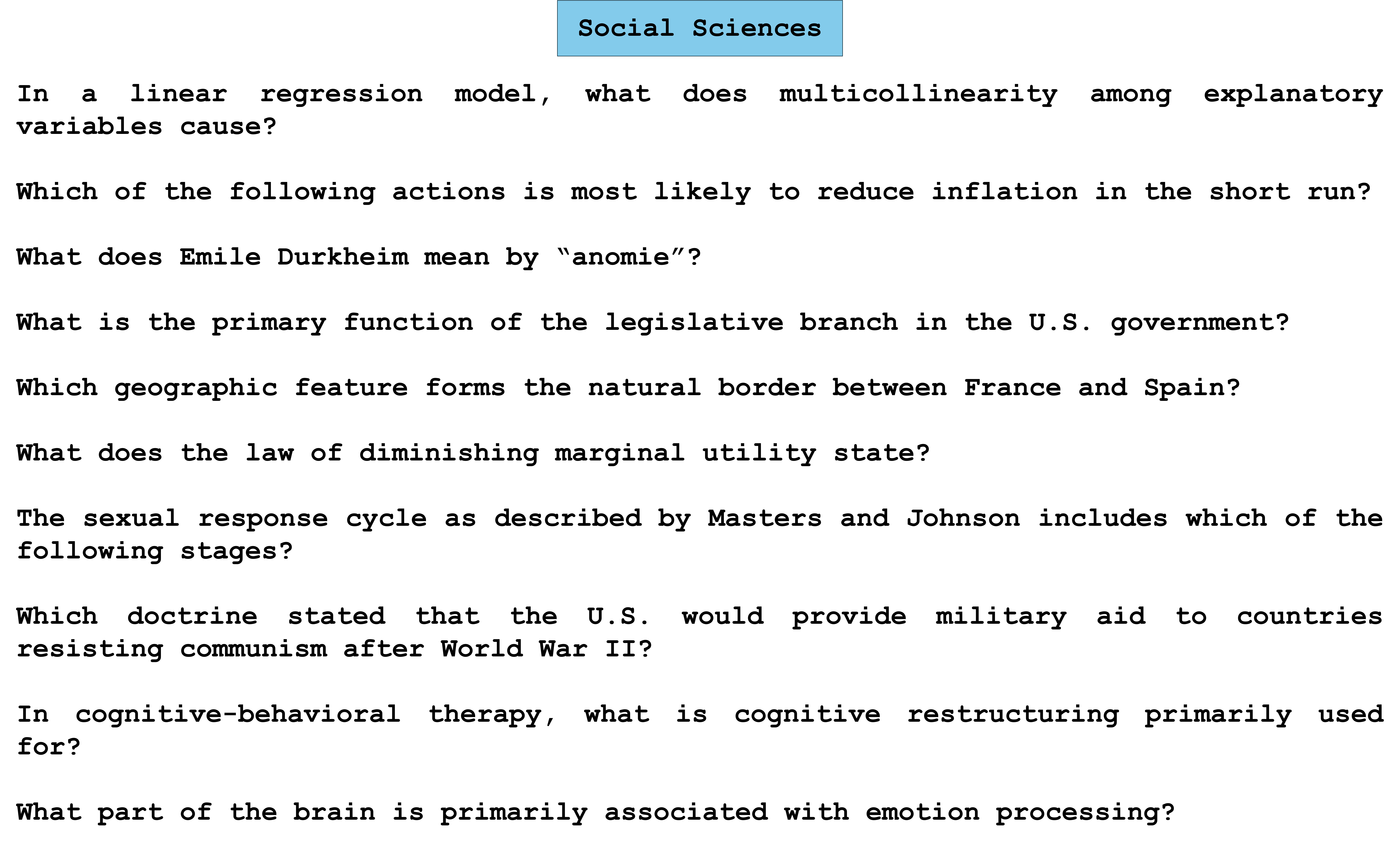}
    \caption{User defined seed sample of Social sciences target domain}
    \label{fig:ss}
\end{figure*}

\begin{figure*}
    \centering
    \includegraphics[width=1\linewidth]{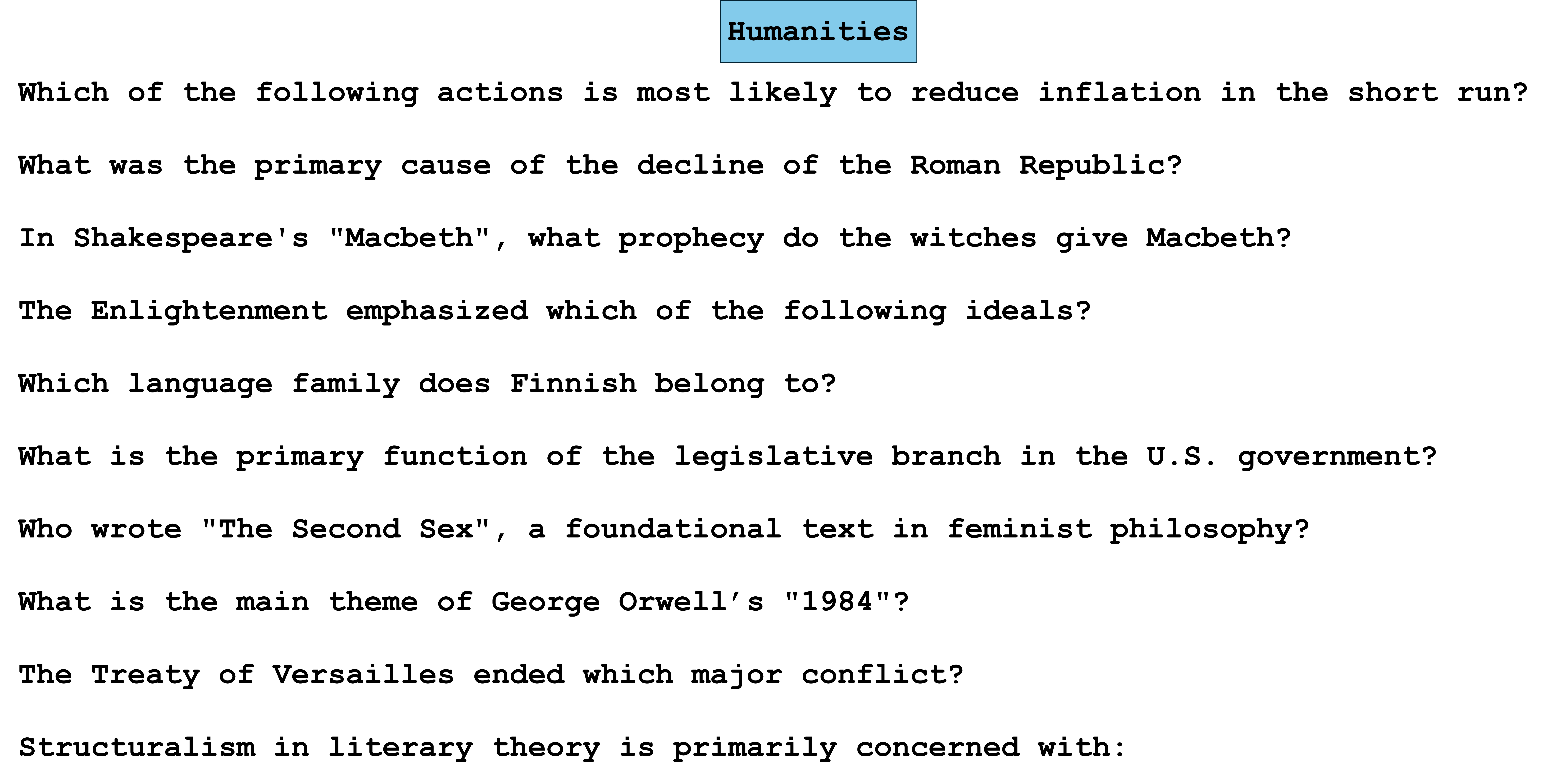}
    \caption{User defined seed sample of Humanities target domain}
    \label{fig:hum}
\end{figure*}

For reproducibility and to clarify the user input required by FineScope, we provide representative seed prompts for each domain; these seeds are the primary human-specified inputs to the pipeline.
Figures~\ref{fig:stem},~\ref{fig:ss}, and ~\ref{fig:hum} present the user-defined seed samples for STEM, Social sciences and Humanities respectively. Based on these seeds, we extracted domain-specific data points from the larger dataset.

\newpage
\bibliography{iclr2026_conference}
\bibliographystyle{unsrt}

\newpage
\end{document}